\documentclass[lettersize,journal]{IEEEtran}
\usepackage{amsmath,amsfonts}
\usepackage{algorithmic}
\usepackage{algorithm}
\usepackage{array}
\usepackage{textcomp}
\usepackage{stfloats}
\usepackage{url}
\usepackage{verbatim}
\usepackage{graphicx}
\usepackage{cite}
\usepackage{multirow,subfigure}
\usepackage{color}
\usepackage{booktabs}
\usepackage{arydshln}
\usepackage{caption}
\usepackage{lipsum}
\usepackage{float}
\usepackage{hyperref,MnSymbol}
\usepackage{xcolor}
\usepackage{fontawesome}
\usepackage{lipsum}

\usepackage{cuted}

\usepackage{subcaption}

\usepackage{kantlipsum}

\usepackage{etoolbox}
\newcommand{\insertfig}{
\captionof*{figure}{}
\setcounter{figure}{0}
\vspace{-2mm}
\centering
    \subfigure[Ground Truth. (24 bpp)]{\includegraphics[width = 0.24\textwidth]{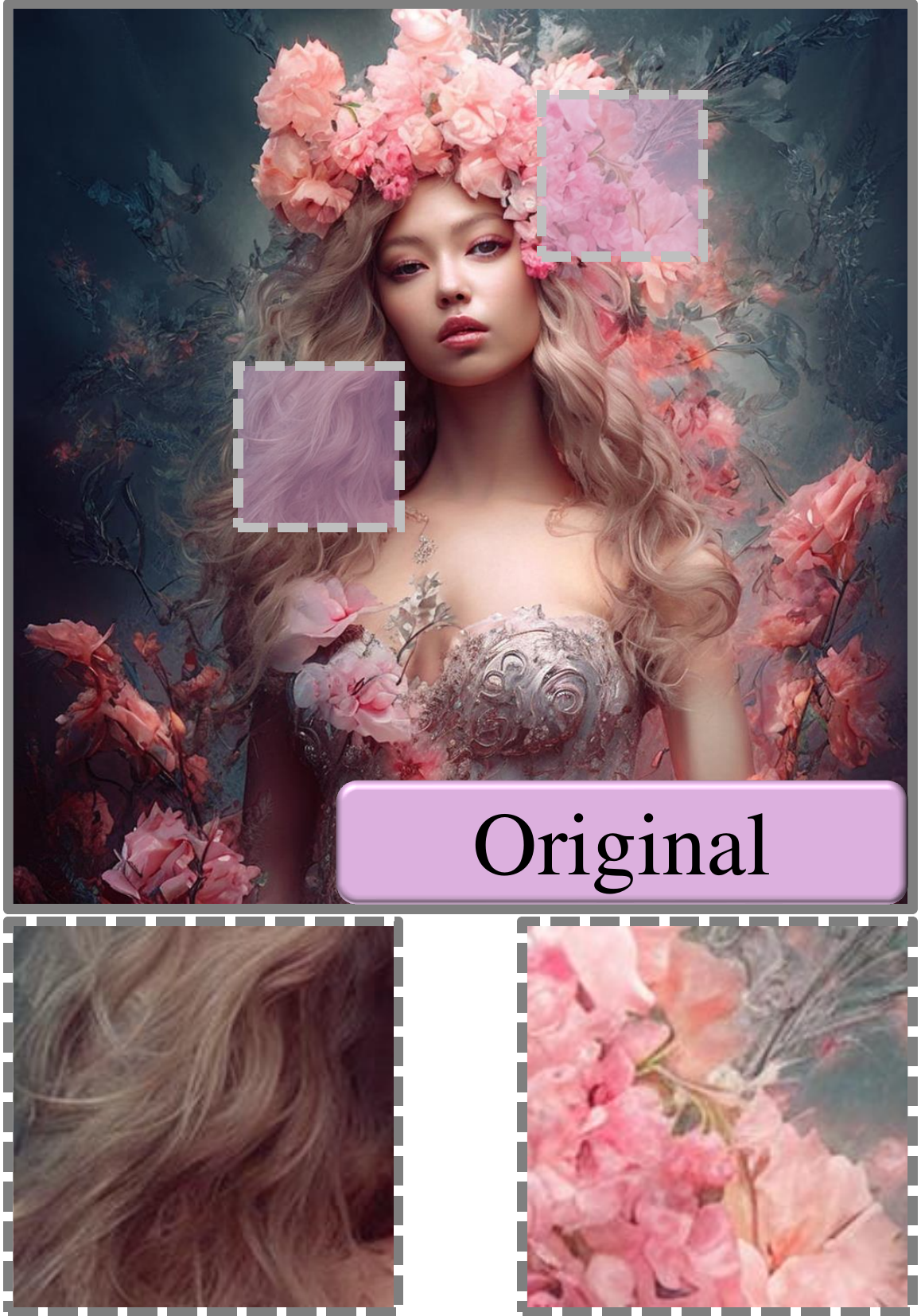}}
    \subfigure[CHENG-2020. (0.057 bpp)]{\includegraphics[width = 0.24\textwidth]{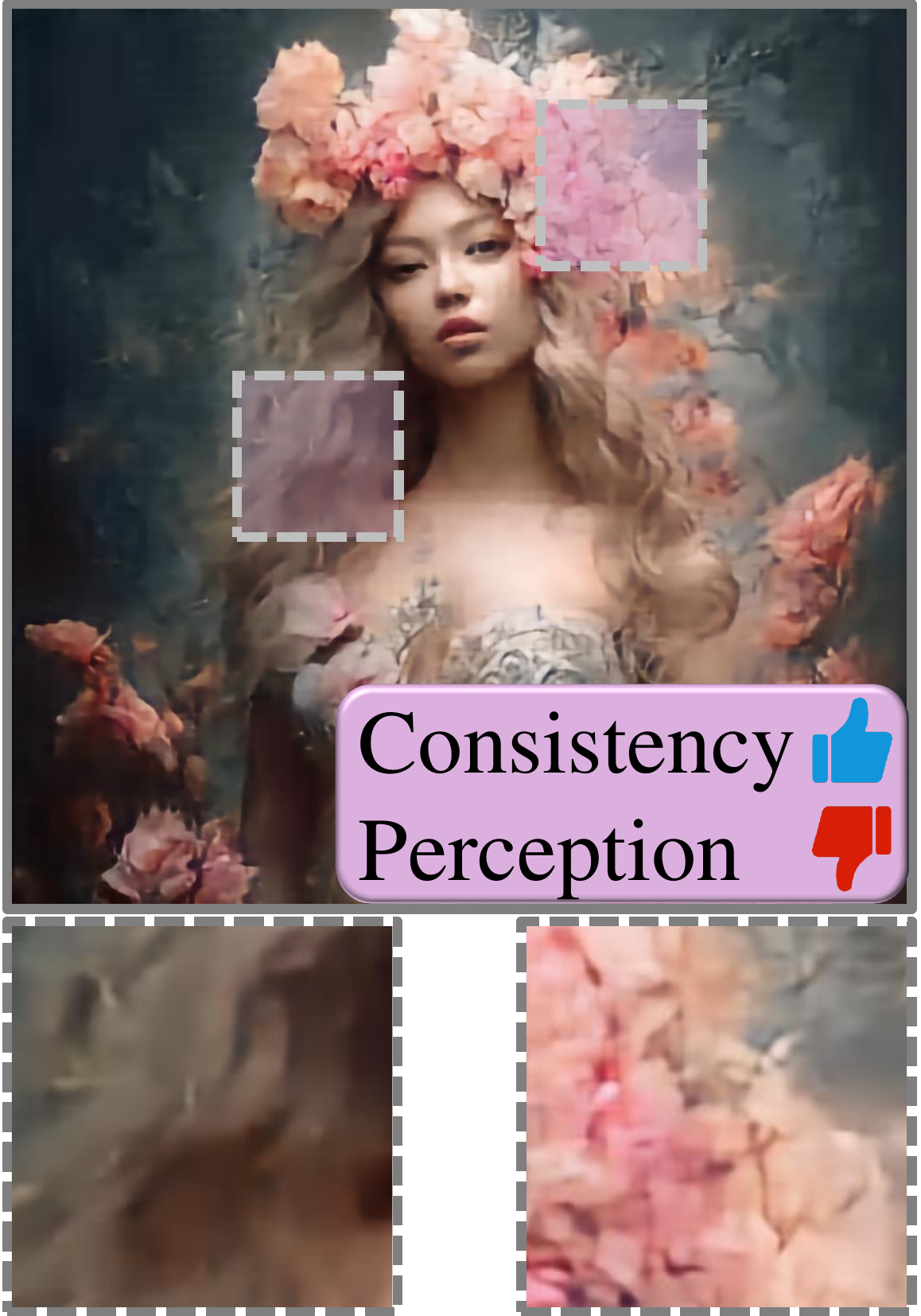}}
    \subfigure[PICS-2023. (0.041 bpp)]{\includegraphics[width = 0.24\textwidth]{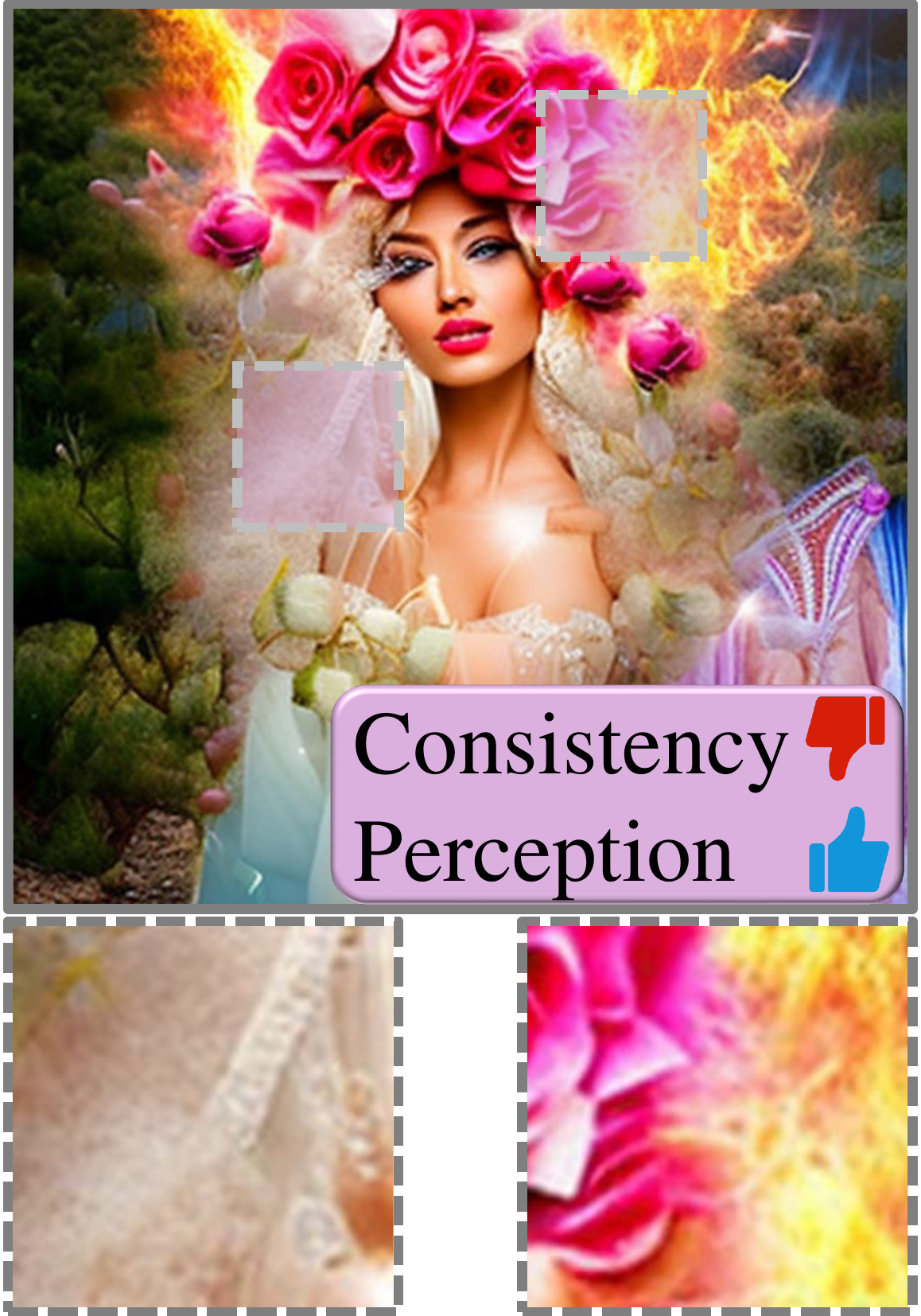}}
    \subfigure[Proposed Method. (0.036 bpp)]{\includegraphics[width = 0.24\textwidth]{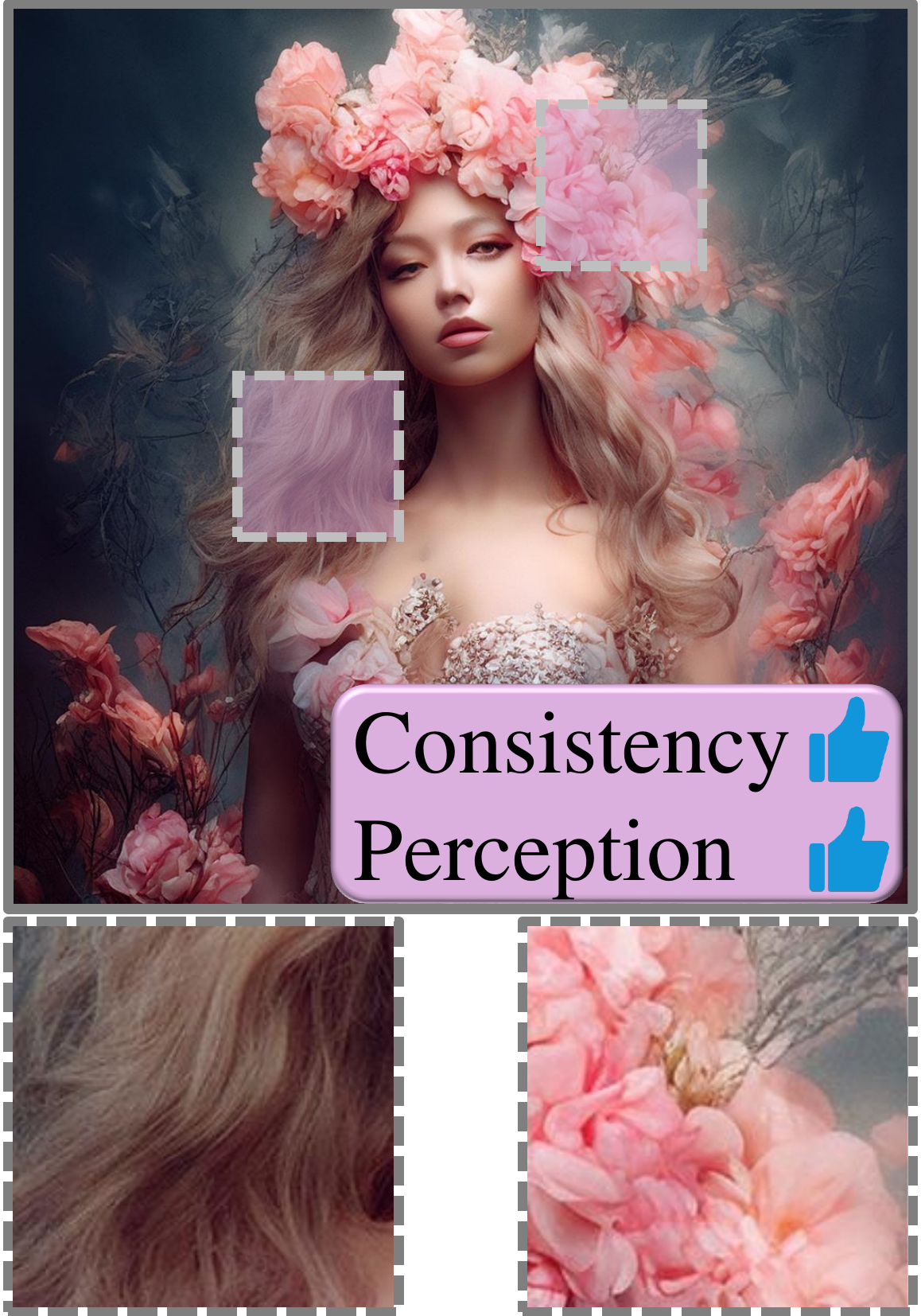}}
\caption{An image compressed by existing algorithm and MISC. In ultra-low bitrate, (b) has no detail which leads to a poor perceptual quality; (c) generates details but inconsistent with ground truth; (d) achieves consistency and perception altogether.}  
\label{fig:spotlight}
\vspace{-2mm}
}

\makeatletter
\apptocmd{\@maketitle}{\centering\insertfig}{}{}
\makeatother

\let\OLDthebibliography\thebibliography
\renewcommand\thebibliography[1]{
  \OLDthebibliography{#1}
  \setlength{\parskip}{0pt}
  \setlength{\itemsep}{0pt plus 0.3ex}
}

\definecolor{cl1}{rgb}{0.56, 0.78, 0.17}
\newcommand{\CLA}[1]{\textcolor{cl1}{#1}}
\begin{document}

\title{MISC: Ultra-low Bitrate Image Semantic Compression Driven by Large Multimodal Model}


\author{Chunyi Li, Guo Lu, Donghui Feng, Haoning Wu, Zicheng Zhang, Xiaohong Liu,\\ Guangtao Zhai,~\IEEEmembership{Senior Member,~IEEE}, Weisi Lin,~\IEEEmembership{Fellow,~IEEE}, Wenjun Zhang,~\IEEEmembership{Fellow,~IEEE}
\thanks{The work was supported by the National Natural Science Foundation of China under Grant 62301310, and by the Shanghai Pujiang Program under Grant 22PJ1406800. Corresponding author: Xiaohong Liu, Guangtao Zhai.}
\thanks{Chunyi Li, Guo Lu, Donghui Feng, Zicheng Zhang, Guangtao Zhai, and Wenjun Zhang are with the Institute of Image Communication and Network Engineering, Shanghai Jiao Tong University, Shanghai 200240, China (email: {lcysyzxdxc,luguo2014,faymek,zzc1998,zhaiguangtao,zhangwenjun}@sjtu.edu.cn)}
\thanks{Xiaohong Liu is with the John Hopcroft Center, Shanghai Jiao Tong University, Shanghai 200240, China (email: xiaohongliu@sjtu.edu.cn)}
\thanks{Haoning Wu and Weisi Lin are with the S-Lab, Nanyang Technological University, Singapore 639798, Singapore (email: haoning001@e.ntu.edu.sg, wslin@ntu.edu.sg)}
\thanks{Manuscript received Mar 11, 2024.}
}

\markboth{Journal of \LaTeX\ Class Files,~Vol.~1, No.~1, Feb~2024}%
{Shell \MakeLowercase{\textit{et al.}}: A Sample Article Using IEEEtran.cls for IEEE Journals}

\newcommand\blfootnote[1]{%
\begingroup
\renewcommand\thefootnote{}\footnote{#1}%
\addtocounter{footnote}{-1}%
\endgroup
}



\maketitle

\begin{abstract}
With the evolution of storage and communication protocols, ultra-low bitrate image compression has become a highly demanding topic. However, all existing compression algorithms must sacrifice either consistency with the ground truth or perceptual quality at ultra-low bitrate. During recent years, the rapid development of the Large Multimodal Model (LMM) has made it possible to balance these two goals. To solve this problem, this paper proposes a method called Multimodal Image Semantic Compression (MISC), which consists of an LMM encoder for extracting the semantic information of the image, a map encoder to locate the region corresponding to the semantic, an image encoder generates an extremely compressed bitstream, and a decoder reconstructs the image based on the above information. Experimental results show that our proposed MISC is suitable for compressing both traditional Natural Sense Images (NSIs) and emerging AI-Generated Images (AIGIs) content. It can achieve optimal consistency and perception results while saving 50\% bitrate, which has strong potential applications in the next generation of storage and communication. The code will be released on \url{https://github.com/lcysyzxdxc/MISC}.
\end{abstract}



\section{Introduction}
\label{sec:intro}
Image compression serves as the foundation for visualizing volumetric signals~\cite{intro:tip-1,intro:tip-2,intro:tip-3}. This technology effectively reduces the storage space and transmission bandwidth required for visual signals without significantly compromising their quality. With the recent advancements in 5G~\cite{my:5G} and 6G~\cite{his:6G}, the integration of numerous embedded devices and Internet-of-Things (IoT) devices into communication protocols has posed challenges due to their limited storage resources and extreme channel conditions. This scenario has made ultra-low bitrate image compression a challenging and demanding research topic, which compresses images to one-thousandth of their original size or even more. To achieve such extreme compression ratios, the focus shifts from low-level fidelity to semantic consistency with the reference image.

However, a trade-off exists between perception and consistency in image compression~\cite{intro:tradeoff-1}. For low bitrate image compression ($< 0.1$ bpp), the compression algorithm provides a rough encoding of the original image, necessitating the decoder to add details. Inadequate detail leads to poor perceptual quality, while excessive detail results in inconsistency with the original image, as illustrated in Fig. \ref{fig:spotlight}. As bitrates decrease further to ultra-low levels ($< 0.024$ bpp, one-thousandth of the original), the conflict between these two objectives becomes even more intensified~\cite{intro:tradeoff-2}.

In recent years, the emergence of AI-Generated Content (AIGC) has revolutionized the field of image compression~\cite{review:GIC,metric:extreme,metric:Semantic}. This paradigm shift enables achieving both perception and consistency at ultra-low bitrates. Leveraging the image understanding capabilities of GPT4-Vision~\cite{intro:gpt4}, Llama~\cite{intro:llama}, and the image generation prowess of Stable Diffusion~\cite{gen:sd,intro:xl} and DALLE~\cite{intro:dalle} series models, images can be compressed into semantic information, facilitating high-quality reconstruction. Additionally, Large Multi-modal Models (LMMs) have altered the content of images to be compressed. Beyond Natural Sense Images (NSIs), AI-Generated Images (AIGIs) have demonstrated significant commercial value by reshaping the creation and marketing of visual content~\cite{intro:aigc}. Given the low-level differences between AIGIs and NSIs (e.g., AI artifacts, texture distribution), the existing compression methods for NSIs may not be suitable for AIGIs, posing an open question of how to compress this distinct image form.
To expand the application of image compression in the AIGC era, we propose Multimodal Image Semantic Compression (MISC) for ultra-low bitrate compression, making the following contributions:

\begin{itemize}
    \item A new paradigm of image compression driven by LMMs. MISC is the pioneering image compression model that integrates LMMs in both the encoder and decoder. This holistic approach will facilitate the extensive utilization of LMMs in image compression applications.
    \item A high-quality AIGI database. We collected 500 high-quality AIGIs generated by today's mainstream Text-to-Image models for future evaluation of the performance of AIGI compression algorithms.
    \item A good balance between consistency and perception. We extensively compared today's mainstream image compression algorithm. Experimental result shows that MISC achieves both satisfactory consistency and perceptual quality for the first time at ultra-low bitrates. 
\end{itemize}

\section{Related Works}
\label{sec:relate}

\begin{table*}[tb]
\centering
\caption{The Structure of existing image compression metric, including en/decoder, format of compressed content, minimum bitrate (Normal/Low/Ultra-low: $>0.1/0.024\sim0.1/<0.024$ bpp), and optimization goal.}
\label{tab:review}
\begin{tabular}{c|l|l|l|l|l|l}
\toprule
Type                                                                       & \multicolumn{1}{c|}{Mertic} & \multicolumn{1}{c|}{Encoder} & \multicolumn{1}{c|}{Content}                                                    & \multicolumn{1}{c|}{Decoder} & \multicolumn{1}{c|}{Min Bitrate} & \multicolumn{1}{c}{Goal} \\ \midrule
Traditional                                                                & Jpeg\cite{metric:tra-jpeg}, Webp\cite{metric:tra-webp}, etc.             & Low-level Processing         & Bitstream                                                                       & Low-level Processing         & Normal                           & Consistency              \\ \hline
\multirow{2}{*}{NIC}                                                       & Mbt2018\cite{metric:nic-mbt}, RDO-PTQ\cite{metric:nic-rdoptq}             & CNN                          & Feature                                                                         & CNN                          & Normal                           & Consistency              \\
    & SReC\cite{metric:nic-srec}, Gao2020\cite{metric:nic-gao}, etc.             & CNN                          & Compressed Image                                                                & CNN+SR                       & Low                              & Consistency              \\ \hline
\multirow{4}{*}{\begin{tabular}[c]{@{}c@{}}GIC\\ (GAN)\end{tabular}}       & Generative-comp\cite{metric:gan-gene}             & CNN                          & Feature                                                                         & Conditional GAN              & Normal                           & Consistency              \\
    & ULCompress\cite{metric:gan-ulc}                  & DWT+GAN                      & Compressed Image                                                                & IDWT+GAN                      & Ultra-low                        & Consistency              \\
    & HiFiC\cite{metric:gan-hific}                       & CNN                          & Feature                                                                         & Conditional GAN              & Normal                           & Perception              \\
    & Multi\cite{metric:gan-multi}/Vari-realism\cite{metric:gan-vari}          & HRRGAN                       & Feature                                                                         & HRRGAN                       & Low                              & Either                  \\ \hline
\multirow{5}{*}{\begin{tabular}[c]{@{}c@{}}GIC\\ (Diffusion)\end{tabular}} & CDC\cite{metric:diff-cdc}                         & Re-DDIM                      & Feature                                                                         & DDIM                         & Normal                           & Consistency              \\
        & Pan2022\cite{metric:diff-pan}                     & Re-SD1.4                     & Feature                                                                         & SD1.4                        & Ultra-low                        & Consistency              \\
       & CMC\cite{metric:diff-cmc}/M-CMC\cite{metric:diff-mcmc}                     & CNN                     & Canny Edge, Text                                                                         &  DDIM+Controlnet                       & Ultra-low                        & Consistency              \\
        & SGC\cite{metric:diff-sgc}                         & DeepLabV3                    & Semantic Map                                                                    & LD                          & Low                              & Perception              \\
        & HFD\cite{metric:diff-hfd}                         & Mean Square Error              & Compressed Image                                                                & SD1.5+Upscaler              & Low                              & Perception              \\
        & Text-Sketch\cite{metric:diff-text}                 & CLIP                         & Canny Edge, Text                                                                & SD2.1+Controlnet            & Ultra-low                        & Perception              \\ \hline
\begin{tabular}[c]{@{}c@{}}GIC\\ (LMM)\end{tabular}                       & MISC (\textbf{proposed})                        & GPT-4 Vision                 & \begin{tabular}[c]{@{}l@{}}Compressed Image, \\ Text, Semantic Map\end{tabular} & SD2.1+Controlnet+Mask                      & Ultra-low                        & Both \\ \bottomrule                  
\end{tabular}
\end{table*}

\subsection{Image Compression Metric}
\label{sec:relate-metric}

Over the past few decades, the evolution of image compression has progressed through four distinct stages as outlined in Table \ref{tab:review}:
\textbf{(\romannumeral1)} The initial stage involved traditional image compression algorithms that utilized low-level visual processing techniques to encode redundancy, such as Macro Blocks (MBs) from H.264 to H.266~\cite{metric:tra-avc,metric:tra-hevc,metric:tra-vvc}. However, these methods primarily focused on pixel-level information, necessitating relatively high bitrates.
\textbf{(\romannumeral2)} Subsequently, with the advancement of deep learning, Neural Image Compression (NIC) emerged as a prominent algorithm. NIC employs end-to-end convolutional neural networks (CNN) in both the encoder and decoder to map the original image to a latent space and restore it. By incorporating Super Resolution (SR) in the decoding process to enhance image details, NIC achieves compression to lower bitrates, but there remains potential for further bitrate reduction.
\textbf{(\romannumeral3)} The introduction of Generative Image Compression (GIC) in 2019 marked a significant development. In contrast to NIC, GIC encodes images within specific constraints to guide the decoder in generating images consistent with the ground truth. Early GIC implementations utilized Generative Adversarial Networks (GANs) as decoders, offering the potential for ultra-low bitrate compression.
\textbf{(\romannumeral4)} After 2022, GAN is gradually replaced by Diffusion, which can be constrained with multimodal information (such as text, edge), encoded by Contrastive Language-Image Pre-Training (CLIP) \cite{model:clip}, and reconstruction models like Denoising Diffusion Implicit Model (DDIM), Latent Diffusion (LD), and Stable Diffusion (SD) have facilitated ultra-low bitrate compression. However, despite the advantages, achieving both consistency and perceptual quality at such low bitrates remains a challenge, necessitating the development of comprehensive metrics that cater to both objectives.

\vspace{-2mm}
\subsection{Image Compression Databases}
\label{sec:relate-database}
Image compression databases play a pivotal role in validating image compression algorithms. For example, Kodak-24 \cite{database:kodak} offers a realistic view of compression effects on authentic images, DIV2K \cite{database:div2k} focuses on high-resolution scenarios, and CLIC-2020 \cite{database:clic} consists of high-quality images with diverse content. However, all the databases above are NSIs, whose characteristics are significantly different from AIGIs. Considering the quality of the existing AIGI database ~\cite{database:agiqa-3k,database:agiqa-1k} is already low, no matter what compression algorithm is used, the resulting image quality is still low.
Therefore, a high-quality AIGI database is needed to measure the performance of image compression algorithms.

\subsection{Evaluation Criteria}
\label{sec:relate-iqa}
Traditionally, the performance of image compression metrics is judged by pixel-level distortion, such as Peak Signal-to-Noise Ratio (PSNR), and Structural Similarity (SSIM) \cite{iqa:ssim}. That's because distortion is relatively small at normal bitrate, and the more consistent the compressed image is with the reference image, the higher the perceptual quality it has. However, as the bitrate decreases, the consistency perception becomes a dilemma \cite{metric:diff-text}, and high-quality images often contain details different from the reference. Therefore, to objectively evaluate the performance of a compression algorithm, both consistency and perception index need to be considered separately.

For consistency, at ultra-low bitrate, pixel-level fidelity metrics are poorly correlated with human subjective perception because they mainly focus on low-level details rather than high-level structures \cite{metric:diff-text}; for perception, traditional LMM image generation works use Fréchet Inception Distance (FID) \cite{iqa:fid} to characterize human preference. However, research shows that subjective human perception does not depend on it, but on signal fidelity and aesthetics ~\cite{add:q-bench,add:q-instruct,add:q-boost,add:q-align,add:q-refine}. Therefore, the semantic similarity between the compressed image and the reference image, and the Image Quality / Aesethic Assessment (IQA/IAA) index ~\cite{his:gms,his:advancing,my:aspect-qoe,my:xgc-vqa,my:cartoon} are more suitable as evaluation criteria of consistency and perception.

\section{Proposed Compression Algorithm}

\begin{figure*}
    \centering
    \includegraphics[width = \textwidth]{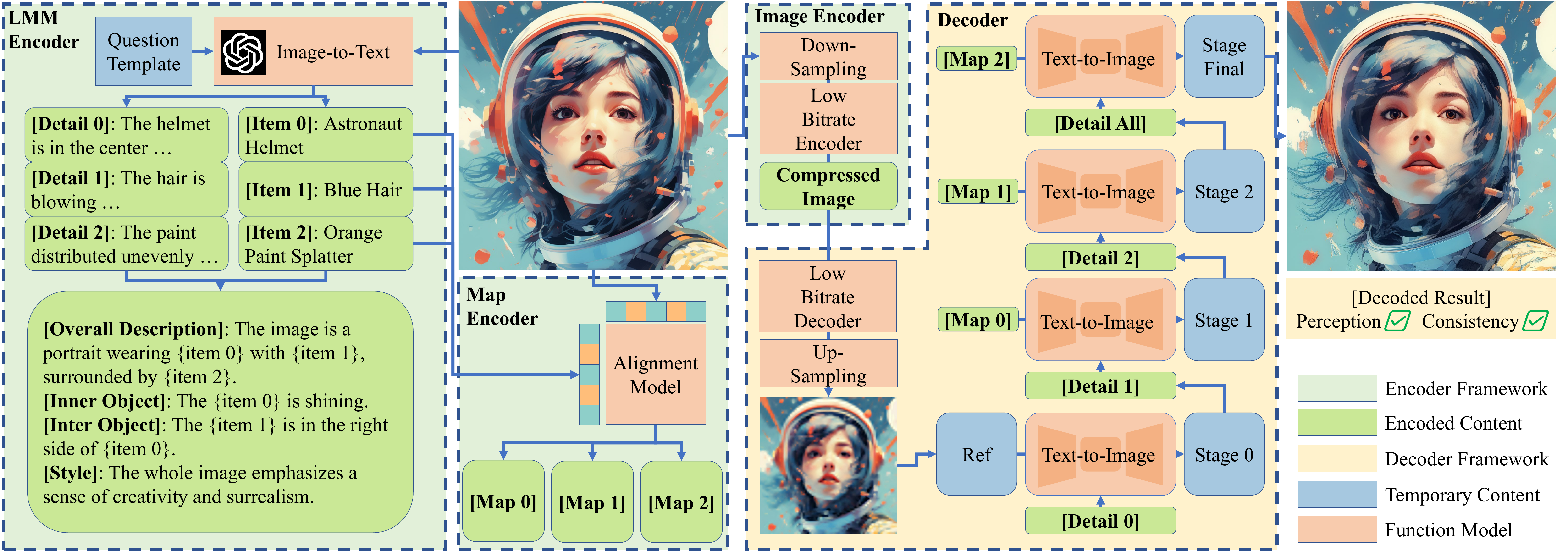}
    \caption{The framework of the MISC model including LMM/map/image encoders and an LMM decoder. The compressed content (noted in \CLA{\textbf{green}}) has an extremely compressed image bitstream lower than 0.024 bpp, a detailed description of a whole image, and items' names, details, and supposed position maps. The decoder controls the diffusion process according to the above content to generate images that simultaneously satisfy high consistency and perceptual quality.}
    \label{fig:framework}
\end{figure*}

\subsection{Framework}
\label{sec:framework}
In this section, we propose the MISC framework for ultra-low bitrate image compression, as shown in Fig. \ref{fig:framework}. Specifically, the framework contains three encoding modules, including an LMM encoder for extracting semantic information of images, a map encoder for annotating regions corresponding to semantic information above, and an image encoder for extreme pixel-level compression, and a decoding module uses the text, map, and image obtained through the above process as constraints to reconstruct the image. Next, we will introduce the three encoders and one decoder module in detail.

\subsection{LMM Encoder}
\label{sec:llm-encode}

\begin{figure}
    \centering
    \subfigure[Frequency Domain Information (traditional compression)]{\includegraphics[width = 0.48\textwidth]{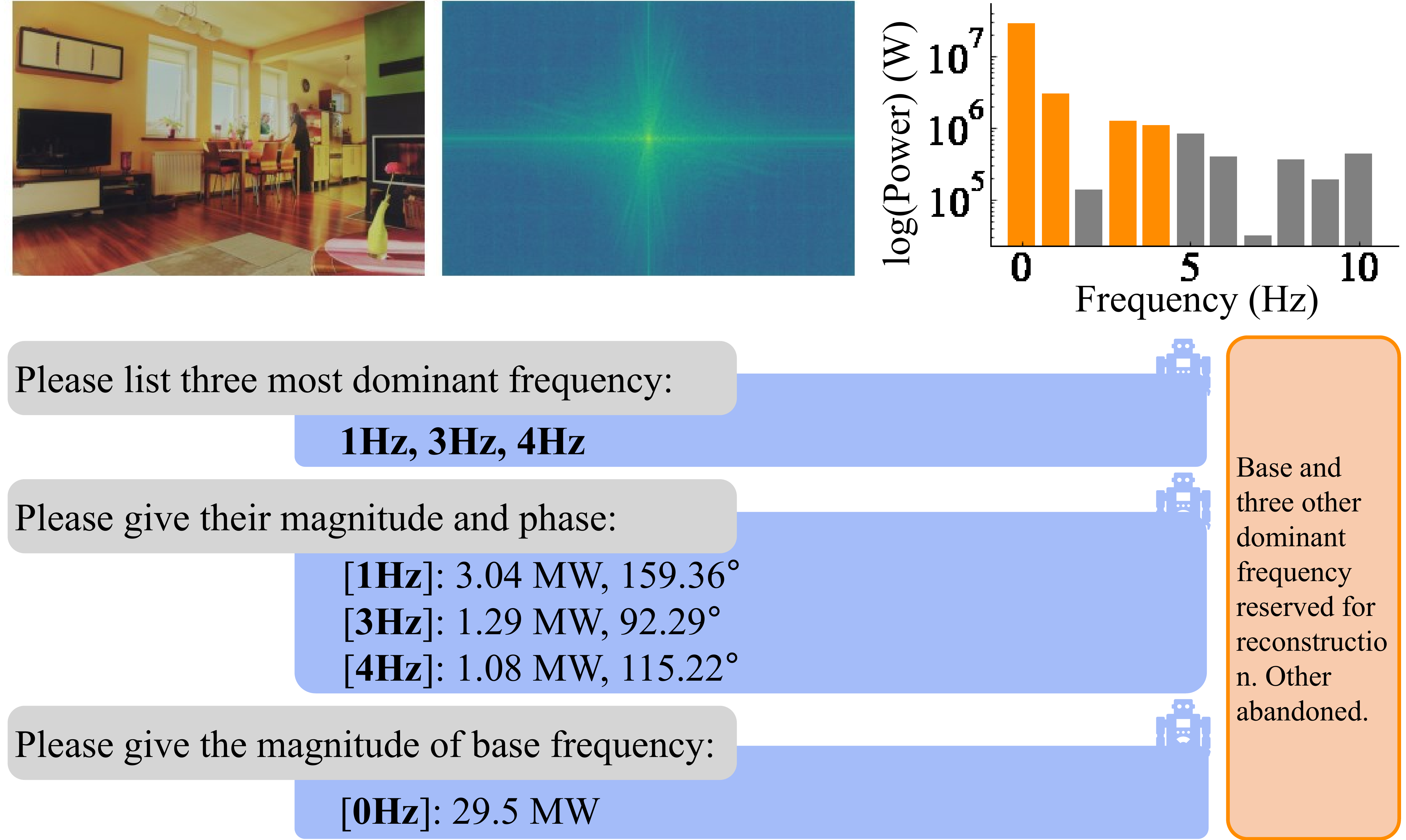}}
    \subfigure[Semantic Domain Information (proposed MISC)]{\includegraphics[width = 0.48\textwidth]{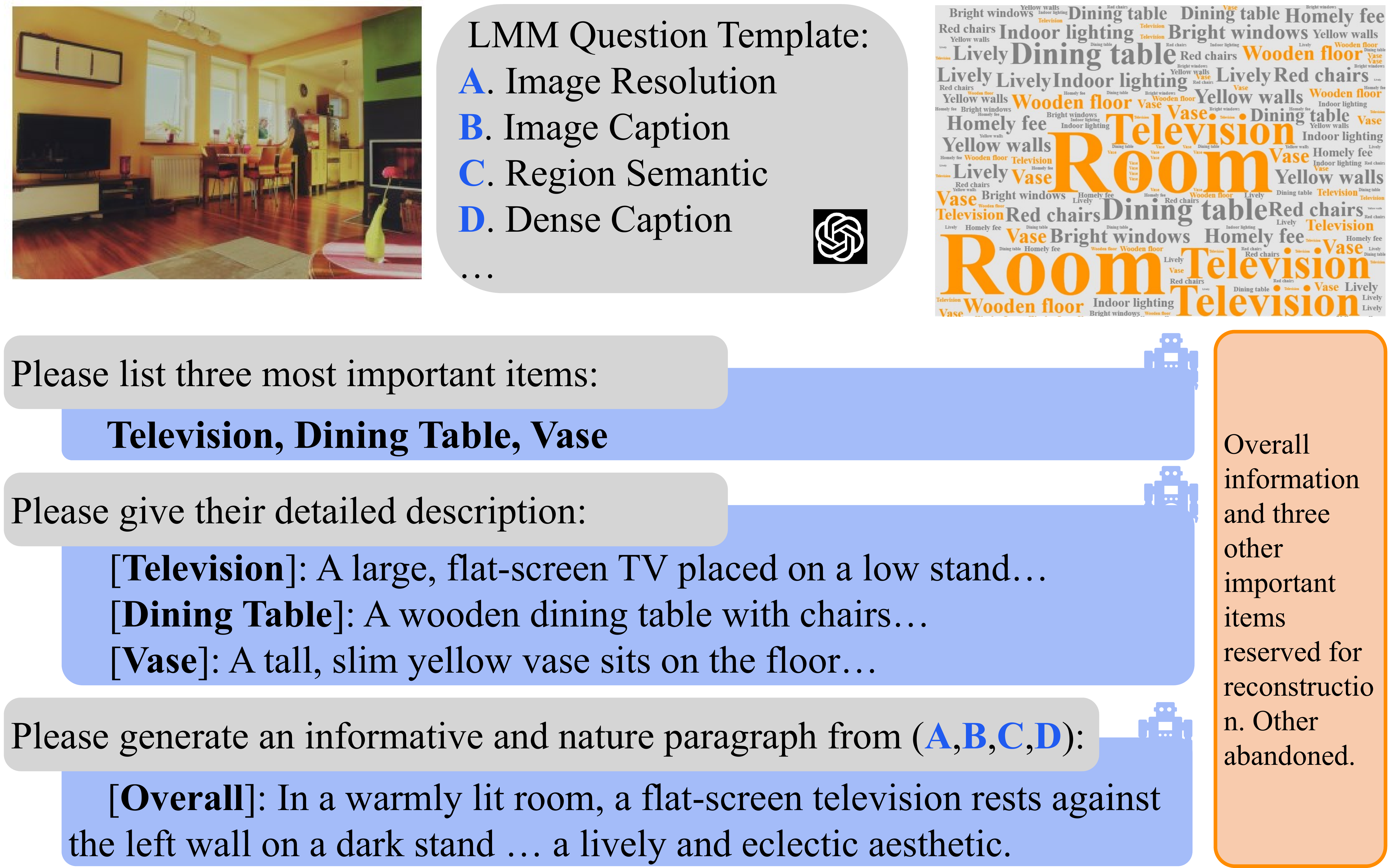}}
    \caption{Comparison of mapping spatial domain into frequency or semantic domain. Both methods compress images by retaining important information and discarding other information.}
    \label{fig:lmm-encode}
    \vspace{-2mm}
\end{figure}

The LMM encoder transforms images into multiple uncorrelated, sparse, weakly dependent semantic variables. It keeps the main variables and discards other variables. Then the decoder can perform an inverse transformation to obtain an image consistent with the original image and save a lot of data space. In recent years, the ability of LMMs to understand and generate images has contributed to more efficient semantic translation. By using Image-to-Text (I2T) and Text-to-Image (T2I) models as encoders and decoders, images can be compressed into more compact semantic information. Unlike Discrete Fourier Transform (DFT) and Discrete Wavelet Transform (DWT), semantically distinct items have lower correlations compared to the frequency domain. Under the same compression performance, the semantic domain can discard more information.
Referring to the spatial-frequency domain, MISC designed the spatial-semantic domain mapping, as shown in Fig. \ref{fig:lmm-encode}. In traditional image compression, the base frequency is usually retained as the overall hue of the image, as well as the amplitude/phase of lower frequencies to represent important details, while other details at higher frequencies are discarded. Similarly, MISC will generate a long description of the image as a whole, as well as a description of some important items in the image, discarding other items. To accurately extract semantic information, MISC asked questions to the most advanced GPT-4 Vision \cite{intro:gpt4} and obtained the following natural language feedback:

\begin{itemize}
    \item $T_n[j]$: Item name ($\leq$ 3 words, $j \in \{0,1, \cdots J-1\}$)
    \item $T_d[j]$: Item detail ($\leq$ 10 words, $j \in \{0,1, \cdots J-1\}$)
    \item $T_{all}$: Detail all ($\approx$ 50 words)
\end{itemize}
where $J$ stands for the number of items. \textbf{[Item name]} is the index of the item and is not directly used for image reconstruction, so it is only represented by no more than three words. \textbf{[Item detail]} includes the shape, color, status, or other attributes. Considering that the description of an object in existing visual question answering \cite{add:q-bench} tasks usually does not exceed 10 words, we set it as an upper limit. \textbf{[Detail all]} is a description of the image as a whole. Past GIC \cite{metric:diff-mcmc} shows that as the number of description words increases, the compression performance gradually increases and reaches a maximum of about 50 words. Considering the data scale of text format, for a $512\times512$ image, a word usually occupies $1\sim2\times10^{-4}$ bpp. To avoid unnecessary overhead, we set 50 as the benchmark.
Assuming in the semantic domain, a few items take up most of the information in the image (just as the low frequencies take up most of the energy),
MISC can set the threshold of the items ${S_{th}}$ referring to the frequency threshold $f_{th}$ of the Macro Block (MB) in H.264\cite{metric:tra-avc} as:
\begin{equation}
    {S_{th}} = \frac{f_{th}}{N_{pix}}{\rm E}({N_{item}}),
    \label{equ:threshold}
\end{equation}
where $(N_{pix},N_{item})$ refer to the number of pixels in an MB, and the number of items in an image with Expectation $\rm E(\cdot)$. According to answers of GPT-4 Vision on Kodak24 \cite{database:kodak} and CLIC2020 \cite{database:clic}, a picture usually contains $10\sim15$ items. From (\ref{equ:threshold}), the optimal threshold $S_{th}=2.81$, so MISC takes $J \leq 3$.

\subsection{Map Encoder}
\label{sec:map-encode}

\begin{figure}[tb]
    \centering
    \subfigure[Item 0: {green bike}]{\includegraphics[width = 0.24\textwidth, height=0.118\textheight]{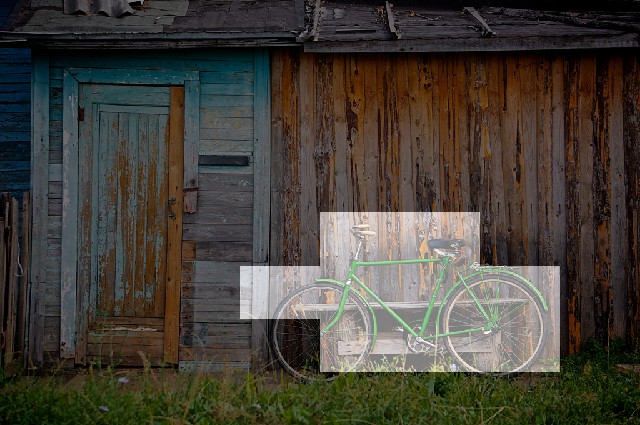}}
    \subfigure[Item 1: {wooden door}]{\includegraphics[width = 0.24\textwidth, height=0.118\textheight]{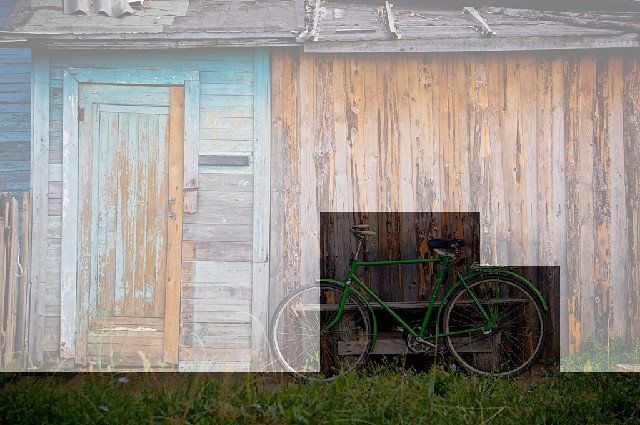}}
    \subfigure[Item 2: {grass}]{\includegraphics[width = 0.24\textwidth, height=0.118\textheight]{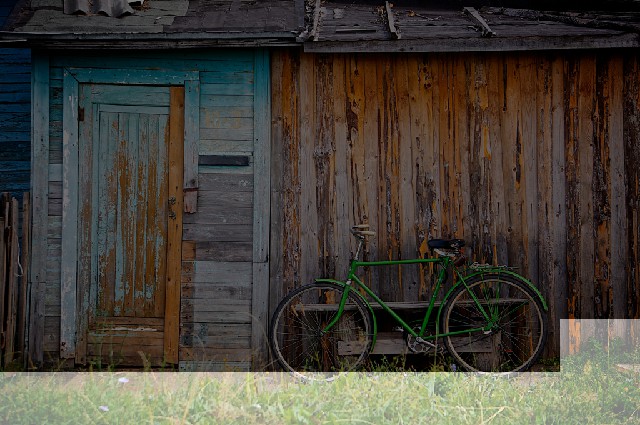}}
    \subfigure[Compressed without map (b)]{\includegraphics[width = 0.24\textwidth, height=0.118\textheight]{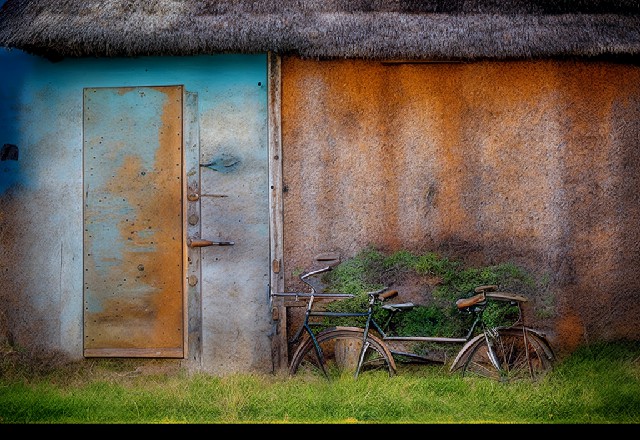}}
    \subfigure[Compressed without map (c)]{\includegraphics[width = 0.24\textwidth, height=0.118\textheight]{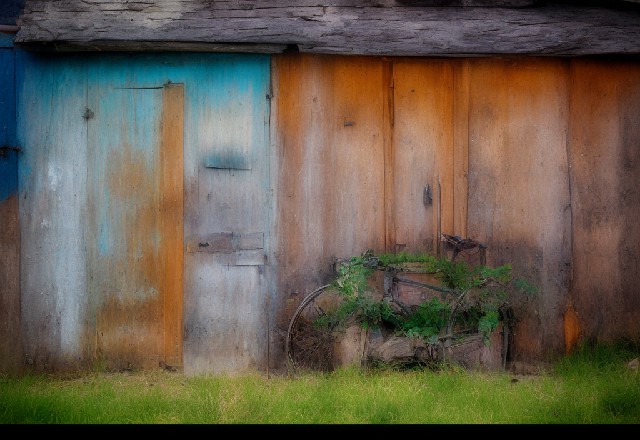}}
    \subfigure[Compressed with all maps]{\includegraphics[width = 0.24\textwidth, height=0.118\textheight]{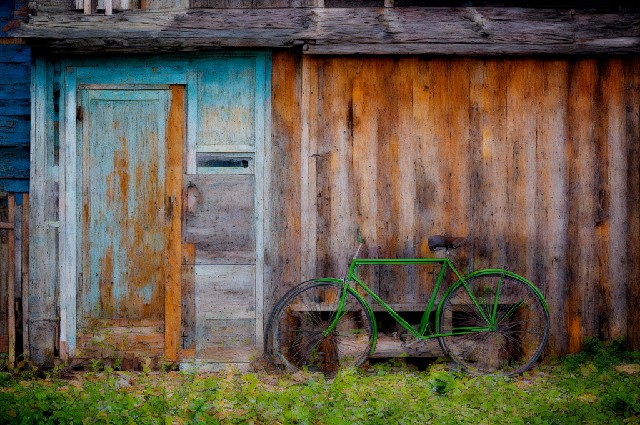}}
    \caption{The positional map of three items, and the decompressed image with/without maps. When (b) and (c) are not used as constraints, `wooden' and `grass' will affect the `bike' region respectively.}
    \label{fig:mask}
    \vspace{-2mm}
\end{figure}

The map encoder acts as an additional module for LMM encoder, by characterizing the spatial relationship between multiple items. This encoder can support a dynamic number of items to balance performance against bitrate. A three-item situation is shown in Fig. \ref{fig:mask} as an example, each map including the following two aspects:
\begin{itemize}
    \item The position of the object itself: For example, [Item 0] is located in the `lower-right' of the entire picture. This word occupies 88 bits, but it is only a rough range; in contrast, an $8\times8$ map only occupies 64 bits and can more accurately specify its location.
    \item The relationship between objects: For example, [Item 2] is to the bottom of [Item 1], and it is difficult to describe the distance between them; in this case, only spatial information can complete this task.
\end{itemize}
Spatial information includes edge and map, of which edge occupies a larger space; considering our ultra-low bitrate requirements, we use maps to mark the corresponding position of each item.
Using the text-image alignment capability of CLIP, we can map the name of the item to the corresponding area of the image. First, multiply the image and text features to obtain a feature matrix $F_{T}$:
\begin{equation}
    {F_{T}}[j] = {\mathcal{C}}_{I}(I) \odot {\mathcal{C}}_{T}({T_n}[j]),
    \label{equ:txt}
\end{equation}
where $I$ denotes the ground truth image and $T_{n}$ means the name of item, $\mathcal{C}_{(I,T)}(\cdot)$ is the CLIP image/text encoder with the length $(L_I,L_T)$. However, on the target detection task, CLIP has a large number of unactivated redundant features, which occupy a large part of the feature space $F_{T}$. Therefore, we adopt CLIP Surgery’s \cite{model:surgery} redundancy elimination method. Input an empty image to get the weight bias $\omega$ of the text encoder, and adjust the $L_T$ dimension on $F_{T}$ to get the redundant feature $F_{R}$:
\begin{equation}
\left\{ \begin{array}{l}
{\omega}[j] = \sum\limits_O{({\mathcal{C}}_{I}({\rm null}) \odot {\mathcal{C}}_{T}({T_n}[j]))}\\
{F_{R}}[j] = {F_{T}}[j] \odot {\omega}[j],
\end{array} \right.
\end{equation}
where $O$ is the number of ${\mathcal{C}}(\cdot)$ output channel. Subtracting them, we have the semantic map $M[i]$ for item $t_n[i]$:
\begin{equation}
\left\{ \begin{array}{l}
M[j] = \sum\limits_O {{\rm{(}}{F_{T}}[j] - {F_{R}}[j])} \\
M[j] = {\rm{Bi}}({\rm{Pool}}(M[j])),
\end{array} \right.
\end{equation}
where ${\rm{Bi}}(\cdot)$ and ${\rm{Pool}}(\cdot)$ represent binarization and pooling functions, which convert the original map into an $8\times8\sim16\times16$ binary matrix to reduce its data size. The data size of maps is extremely small ($<10^{-3}$ bpp), but it greatly makes up for the shortcomings of the LMM encoder. As shown in Fig. \ref{fig:mask}, if the map of [Item 1] or [Item 2] is missing, [Item 0] `bike' will become `wooden' or grow `grass' during the decoding process. Only by using map constraints for each item can the desired content be generated at the exact location.

\subsection{Image Encoder}
\label{sec:image-encode}
The image encoder provides a reference for the decoder through an extremely compressed image. By combining it with the two encoders above, the reconstruction result from the decoder can have a satisfying perceptual quality while avoiding major defects in consistency with the original image.
This image only includes the rough outline of the original image, which can greatly improve the consistency score. At the same time, although the perceptual quality of this image is poor, the decoder reconstruction process will add certain details, and the perception score of the final decoded image is still acceptable.
To achieve a balance between consistency and perception, we followed this idea under ultra-low bitrate and compressed the image into bitstream $B$:
\begin{equation}
B = \left\{ {\begin{array}{*{20}{c}}
\begin{array}{l}
{\rm{En}}(S_x^ - (I))\\
{\rm{Qu}}(S_x^ - (I))
\end{array}&\begin{array}{l}
x \le {\rm{size}}(I)/{{\rm{En}}_{th}}\\
x > {\rm{size}}(I)/{{\rm{En}}_{th}},
\end{array}
\end{array}} \right.
\label{equ:image-encode}
\end{equation}
where ${\rm{En}}(\cdot)$ and ${\rm{Qu}}(\cdot)$ are the CNN encoder and the quantization function. $S_x^ {(+,-)} $ stands for up/down-sampling. When the down-sampling rate $x$ is low, we can directly use the existing low bitrate NIC/GIC encoder in TABLE \ref{tab:review} to output the compressed bit stream; when $x$ becomes high, the image size is lower than the minimum input threshold ${{\rm{En}}_{th}}$ of the encoder. Thus, the value of each pixel is directly quantized, and the image is treated as a matrix and output as bitstream. Therefore, by inputting the bitstream together with maps and texts into the decoder, a high-consistency and high-perceptual quality reconstruction process can be achieved.

\subsection{Decoder}
\label{sec:decode}
The decoding process is based on the following information provided by the encoders, including a bitstream (representing an extremely compressed image), a detailed description of the overall image, and several name-detail-map (NDM) groups characterizing items in the image. Based on the semantic domain analysis in Sec. \ref{sec:llm-encode}, we take the number of group index $j$ as (\romannumeral1) $J=3$ to help image reconstruction, as reference information is limited at ultra-low bitrate; (\romannumeral2) $J=0$ to save bitrate, as relatively more information is contained at higher bitrate. In the first step, the decoder will perform the opposite operation of (\ref{equ:image-encode}) to obtain the reference image $I_{ref}$:
\begin{equation}
I_{ref} = \left\{ {\begin{array}{*{20}{c}}
\begin{array}{l}
S_x^ + {\rm{(De}}(B))\\
S_x^ + {\rm{(BIC}}(B))
\end{array}&\begin{array}{l}
x \le {\rm{size}}(I)/{\rm{E}}{{\rm{n}}_{th}}\\
x > {\rm{size}}(I)/{\rm{E}}{{\rm{n}}_{th}},
\end{array}
\end{array}} \right.
\end{equation}
where ${\rm{De}}(\cdot)$ and ${\rm{BIC}}(\cdot)$ are the CNN decoder and the bi-cubic interpolation.
Then we extract the probability density ${z}$ from text $T_d[j]$, and conduct diffusion progress referring to ${z}$ with output $S[j]$:
\begin{equation}
\left\{ \begin{array}{l}
z = {\rm{QKV}}({T_d}[j])\\
S[j] = {\cal D}_n^z({\cal D}_{n - 1}^z \cdots {\cal D}_1^z(S[j-1]))\\
S[j] = S[j]M[j] + S[j-1](1 - M[j]),
\end{array} \right.
\label{equ:diff}
\end{equation}
Where $\rm QKV(\cdot)$ represents the multi-head attention and ${\cal D}_n^z$ denotes the diffusion operation at the $n$-th iteration.
To reconstruct an image that aligns with both the reference image $I_{ref}$ and the NDM groups, we utilize the weight of DiffBIR \cite{model:diffbir} in the diffusion model ${\cal D}$ to integrate the image reference and text description comprehensively.
For the diffusion on [Item j], we only update the target region in $S[j]$ based on $M[j]$, while keeping other regions unchanged as the previous state $S[j-1]$. The original state is set as $S[-1]=I_{ref}$. After iterating this process for $J$ times, referring to the descriptive text $T_{all}$ of the entire image, we can enhance details in $S[J-1]$ and obtain the final decoding result $S[final]$:
\begin{equation}
\left\{ \begin{array}{l}
z = {\rm{QKV}}({T_{all}+T_{aes}})\\
S[final] = {\cal D}_{\hat{n}}^z({\cal D}_{\hat{n} - 1}^z \cdots {\cal D}_1^z(S[2])),
\end{array} \right.
\end{equation}
where $\hat{n}$ is $4\sim8$ times the value of $n$ and $T_{aes}$ represent prompts that guide the high-quality generation (e.g., `hyper detail', `masterpiece', `4K'). This step plays a crucial role in the decoding process as it is responsible for ensuring consistency and enhancing perception scores. To prioritize this step, more iterations are allocated to it, and $T_{aes}$ is fed into the QKV mechanism along with $T_{all}$. This allows the decoder to effectively balance these two conflicting objectives, particularly at ultra-low bitrates.

\section{Proposed Database}
\subsection{Data Collection}
\label{sec:collect}
\begin{table}[tb]
\centering
\caption{Existing Text-to-Image generative model. The database named AIGI-SCD is constructed by models with the best popularity (download times) and quality (overall IQA score, normalized average from ClipIQA$\uparrow$ \cite{quality:CLIPIQA}, DBCNN$\uparrow$ \cite{quality:DBCNN}, LIQE$\uparrow$ \cite{quality:LIQE}, NIQE$\downarrow$ \cite{quality:NIQE}). The top three popularity/quality models are marked in {\color[HTML]{FF0000}\textbf{red}}.}
\label{tab:T2I}
\begin{tabular}{l|r|rr}
\toprule
\multicolumn{1}{c|}{\multirow{2}{*}{Metric}} & Popularity                      & \multicolumn{2}{c}{Quality}                                           \\ \cline{2-4} 
\multicolumn{1}{c|}{}                       & \multicolumn{1}{c|}{Download}  & \multicolumn{1}{c}{Score($\uparrow,\uparrow,\uparrow,\downarrow$)}                 & \multicolumn{1}{c}{Average} \\ \hline
SDXL\cite{intro:xl}                                        & {\color[HTML]{FF0000}\textbf{4,400K}}                                             & (0.63, 0.64, 3.20, 4.27) & 0.69                         \\
SD1.5\cite{gen:sd}                                       & {\color[HTML]{FF0000}\textbf{3,780K}}                                             & (0.66, 0.55, 3.21, 4.56) & -0.06                         \\
SD1.4\cite{gen:sd}                                       & {\color[HTML]{FF0000}\textbf{2,640K}}                                               & (0.66, 0.57, 3.24, 4.37) & 0.83                         \\
SDXL-Turbo\cite{gen:turbo}                                  & 799K                                                & (0.59, 0.64, 3.97, 4.41) & 1.28                         \\
Midjourney\cite{gen:MJ}                                  & 248K                                                & (0.71, 0.62, 4.09, 5.05) & 1.29                         \\
DALLE-2\cite{intro:dalle}                                     & 240K                                                & (0.69, 0.48, 2.54, 5.10) & -2.55                         \\
SSD\cite{gen:ssd-1b}                                         & 236K                                                & (0.66, 0.66, 3.79, 5.02) & 0.62                         \\
Playground\cite{gen:Playground}                                  & 132K                                                & (0.70, 0.68, 3.66, 4.92) & {\color[HTML]{FF0000}\textbf{1.57}}                        \\
Deramlike\cite{gen:dream}                                   & 105K                                                & (0.68, 0.61, 3.88, 4.62) & {\color[HTML]{FF0000}\textbf{1.68}}                        \\
Pixart\cite{gen:pixart}                                      & 53K                                                 & (0.72, 0.62, 3.79, 4.47) & {\color[HTML]{FF0000}\textbf{2.75}}                        \\
IF\cite{gen:IF}                                          & 33K                                                 & (0.68, 0.54, 2.85, 5.00) & -1.31                        \\ \bottomrule
\end{tabular}
\end{table}
Beyond NSIs, to compare the performance of image compression algorithms on AIGIs, we construct an AIGI Semantic Compression Database (AIGI-SCD). For a fair comparison, its data needs to represent most of the existing AIGIs, and should not have distortion (otherwise it will overlap with compression distortion, affecting the evaluation of the compression algorithm). Therefore, we conducted a detailed survey of today's mainstream T2I models and selected the most popular and highest-quality models for image generation. According to the huggingface website\footnote{Data collected in January 2024. Considering that DALLE 2 and Midjourney are closed source models, their download times are replaced by OpenDALLE and Openjourney.}, we listed 11 common AIGI models, used their download times as popularity indicators, and generated 500 images for each; then, we used four IQA indicators to evaluate the images generated by each model. Quality scores $S$ were collected four times, and each column in TABLE \ref{tab:T2I} is normalized to $-1\sim1$ as:
\begin{equation}
\overline{{{S_{p,q}}}}=2\cdot \frac{S_{p,q}-{\rm min}{(S_{\cdot,q})}}{{\rm max}(S_{\cdot,q })-{\rm min}(S_{\cdot,q })}-1,
\label{equ:avg1}
\end{equation}
where $p$ represents the metric column in TABLE \ref{tab:T2I} while indicators $q \in$ (ClipIQA, DBCNN, LIQE, NIQE)~\cite{quality:CLIPIQA,quality:DBCNN,quality:LIQE,quality:NIQE} in the quality column. Then the average score $A$ for each row's metric can be formulated below:
\begin{equation}
A_{p}=\sum_{q}(\pm 1)\cdot \overline{{{S_{p,q}}}},
\label{equ:avg2}
\end{equation}
where we use $+1$ for upper-better and $-1$ for lower-better metric. According to Table \ref{tab:T2I}, the most widely used SD series models ~\cite{gen:sd,intro:xl} are representatives of AIGI and need to be included in this database. Meanwhile, the latest Pixart\cite{gen:pixart}, Playground\cite{gen:Playground}, and Dreamlike\cite{gen:dream} models generate the highest quality results, thus suitable for compression tasks as the raw images. Considering the high similarity between SD1.5 and 1.4, we select SDXL and SD1.5 to generate mainstream AIGIs and use the above three high-quality generation models to characterize emerging AIGIs. Referring to the data scale and division of CLIC2020\cite{database:clic}, the AIGI-SCD contains 100 images from each of the five generative models, 450 are for training and 50 for testing.
\begin{figure}[tb]
    \centering
    \subfigure[Kodak\cite{database:kodak}]{\includegraphics[width = 0.24\textwidth]{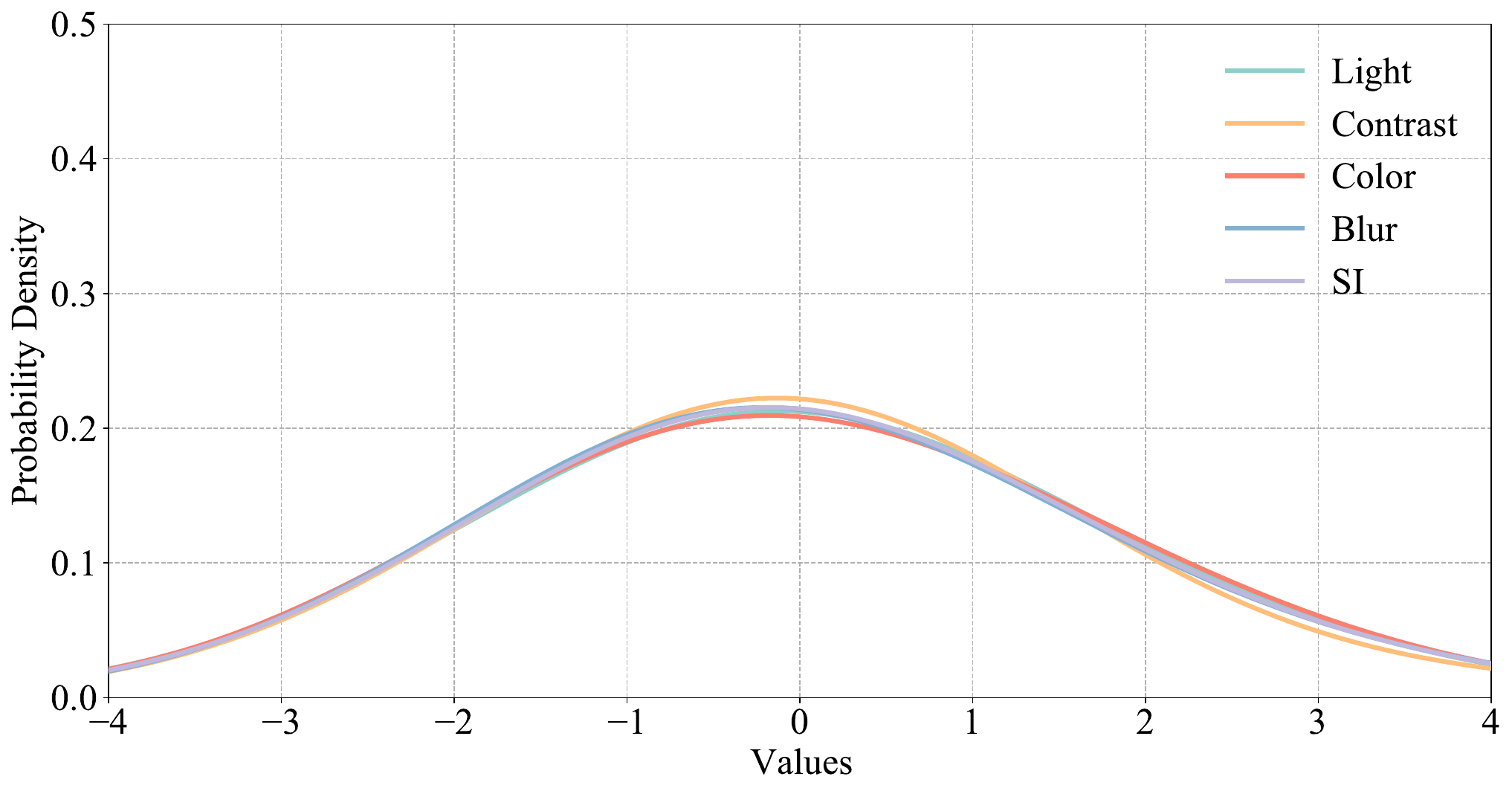}}
    \subfigure[CLIC2020\cite{database:clic}]{\includegraphics[width = 0.24\textwidth]{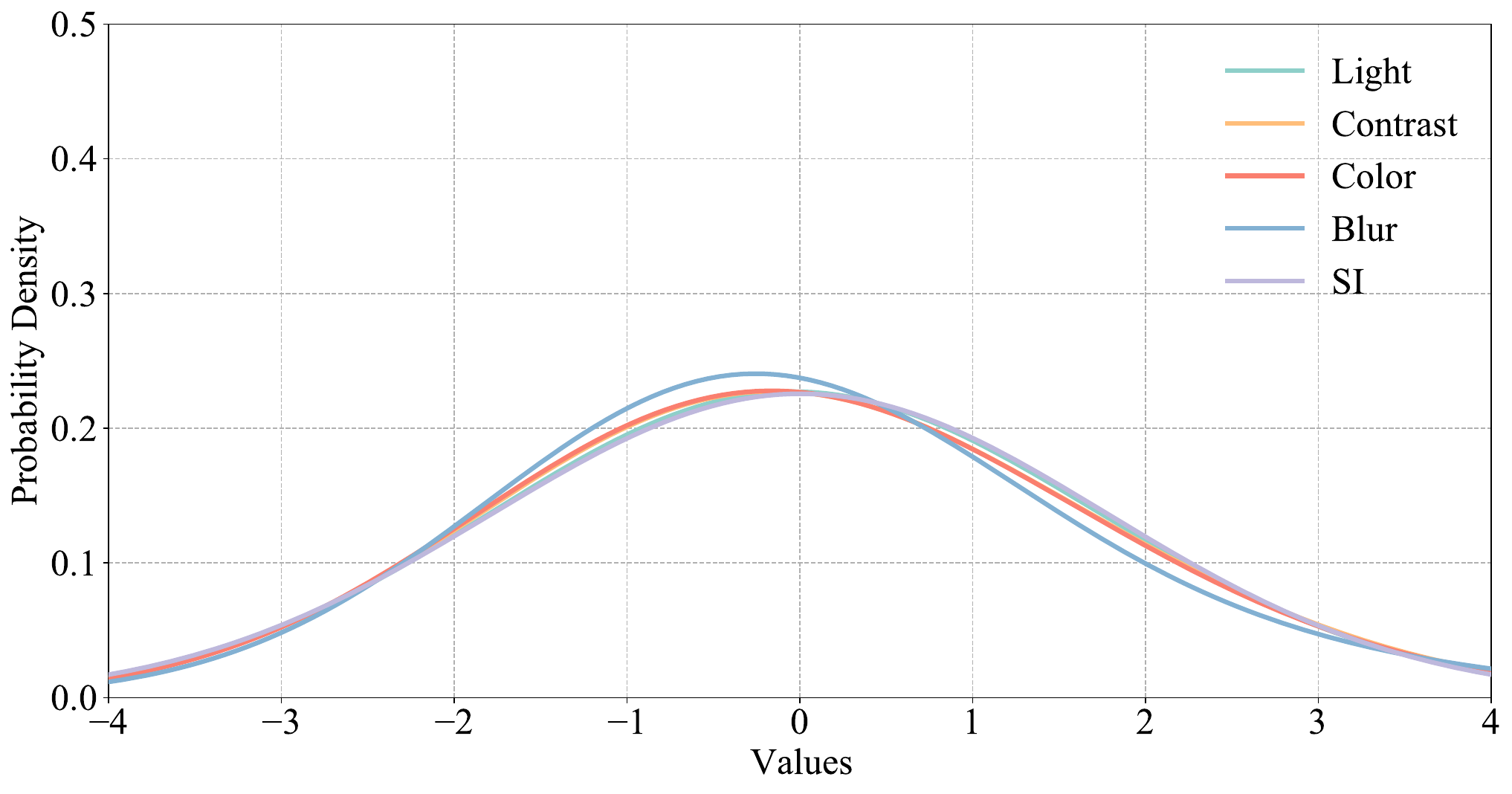}}
    \subfigure[Tecnick\cite{database:tecnick}]{\includegraphics[width = 0.24\textwidth]{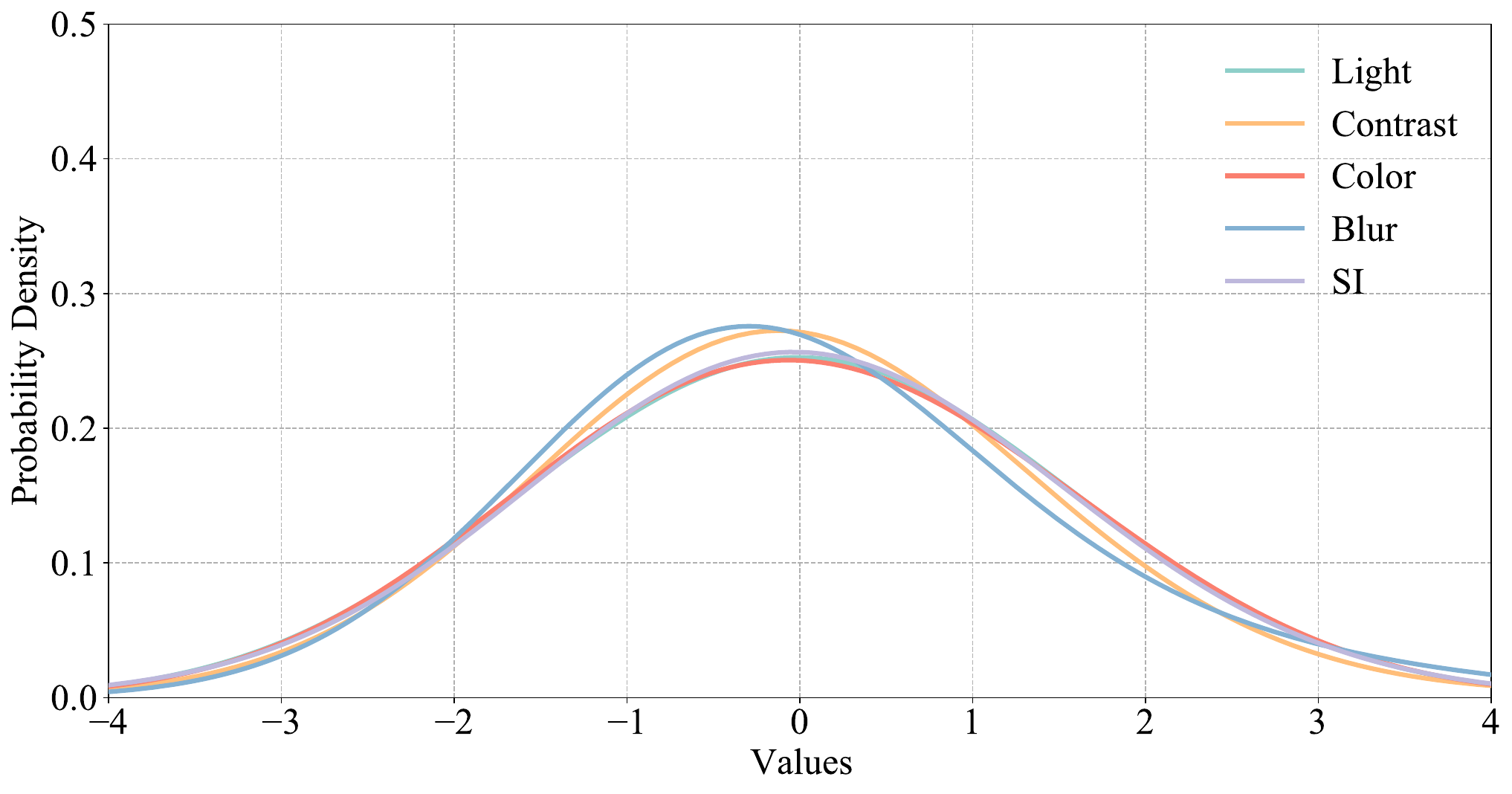}}
    \subfigure[AIGI-SCD]{\includegraphics[width = 0.24\textwidth]{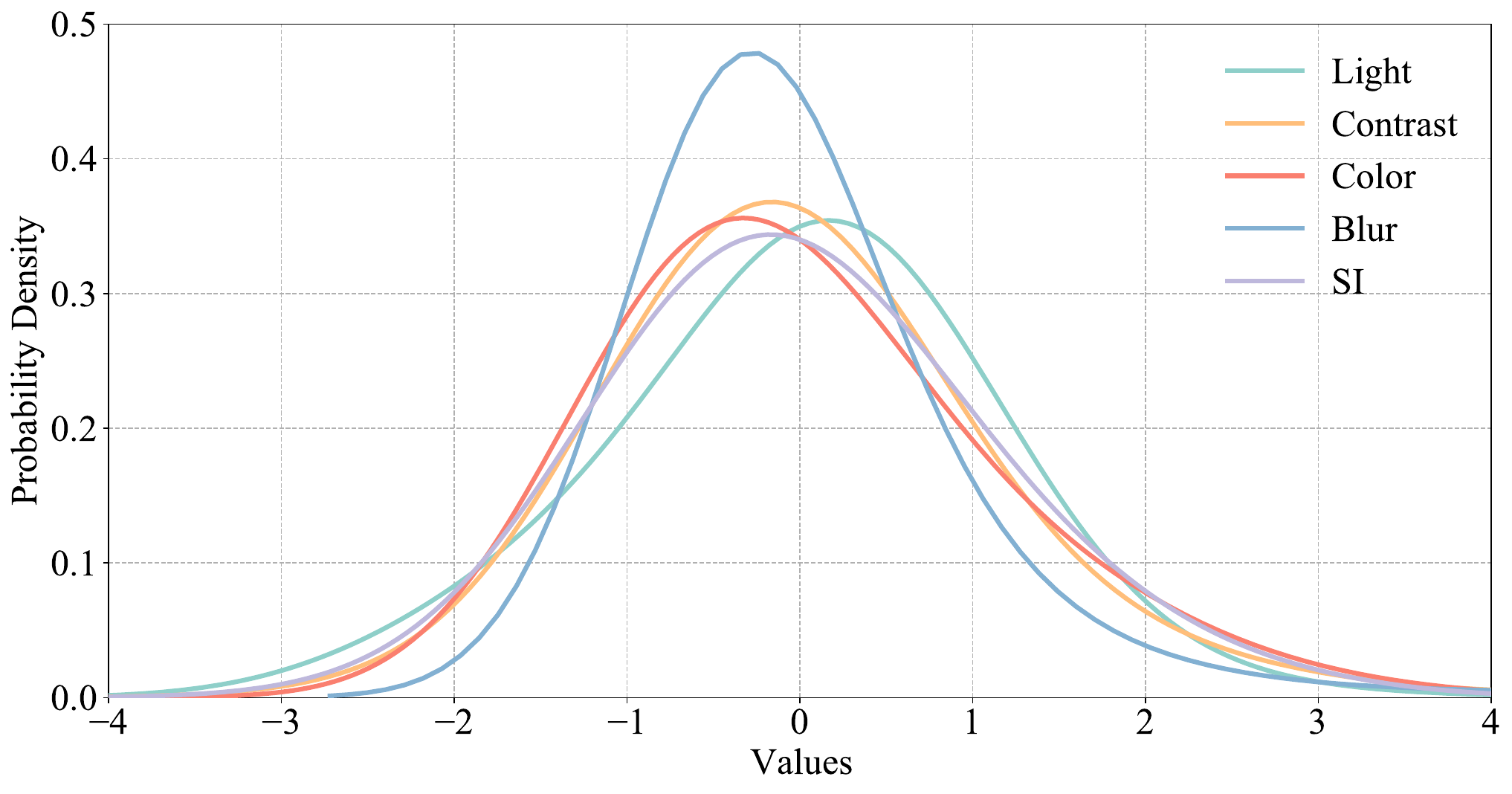}}
    \caption{The normalized probability distributions of the low-level attributes. The distributions include NSIs in the Kodak24 \cite{database:kodak}, CLIC2020 \cite{database:clic}, Tecnick \cite{database:tecnick}, and the proposed AIGI-SCD database. The AIGIs have a sharper distribution and more common blur.}
    \label{fig:curve}
\end{figure}
\begin{figure}[tb]
    \centering
    \vspace{-2mm}
    \includegraphics[width = 0.45\textwidth]{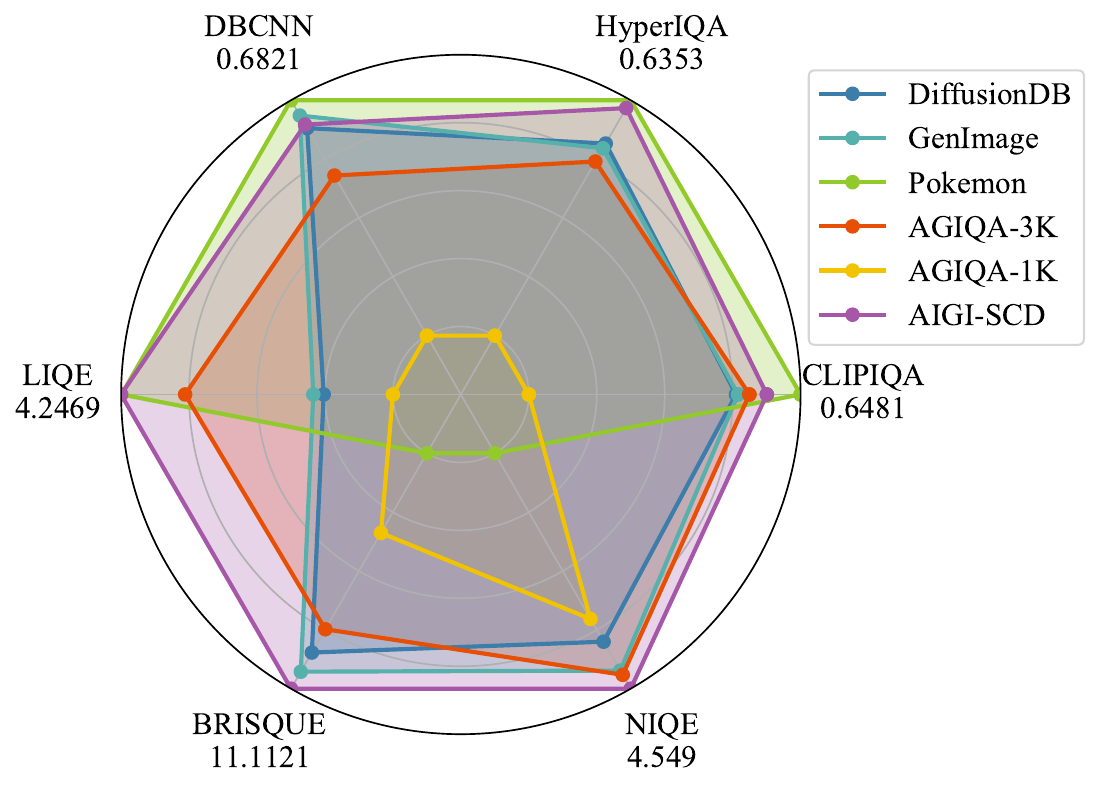}
    \caption{Quality score comparison of the AIGI database. All existing AIGI databases have flaws in at least one quality indicator, while six quality scores of AIGI-SCD are all satisfactory, making them suitable for image compression tasks.}
    \label{fig:database}
    \vspace{-3mm}
\end{figure}
\subsection{Data Analysis}
This chapter will conduct a detailed analysis of the properties of the MISC-AIGI database. By comparing it with existing databases, we prove the importance of this database on AIGI compression tasks.
For the existing NSI databases, to analyze their statistical differences against AIGIs, we use the distribution of five compression-related attributes for comparison, including light, contrast, color, blur, and Spatial Information (SI). Detailed descriptions of these attributes are defined in \cite{add:define}. We selected three NSI compression databases, namely Kodak24 \cite{database:kodak}, CLIC2020 \cite{database:clic}, and Tecnick \cite{database:tecnick} for comparison. As shown in Fig. \ref{fig:curve}, the attributes of NSI and AIGI are both normally distributed, indicating that AIGI-SCD has the same diversity as the traditional NSIs. However, compared to NSIs captured in the wild, the distribution curve of AIGIs is sharper. For example, AIGI is not affected by overexposure distortion in photography, so brightness and contrast are relatively stable. In addition, as AIGIs sometimes have limited iterations during the generation process, the image may suffer from some blurry regions. Thus, the center of the blur curve is biased to the left. These differences indicate some compression mechanisms for NSIs, such as the equalization of brightness and chrominance, and the enhancement of blurry regions may not be effective for AIGIs. In all, for a compression algorithm trained with the traditional \textit{NSI database}, its compression performance on \textit{AIGIs} is likely to be unsatisfactory due to their \textit{different low-level attributes}.
\begin{table*}[tbph]
\centering
\caption{Performance of the state-of-art compression metrics and MISC-1/3 levels, validated on NSI database CLIC2020\cite{database:clic}, and AIGI database AIGI-SCD we constructed. The quality is evaluated by two consistency (LPIPS$\downarrow$ \cite{iqa:lpips}, ClipSIM$\uparrow$ \cite{model:clip}) and two perception (NIQE$\downarrow$ \cite{quality:NIQE}, ClipIQA$\uparrow$ \cite{quality:CLIPIQA}) indicators. A normalized average of those four indicators from equation (\ref{equ:avg1}) (\ref{equ:avg2}) is provided as the overall evaluation of consistency and perception together. [Key: {\color[HTML]{FF0000} \textbf{Best}}; {\color[HTML]{4472C4}\textbf{Second Best}}].}
\label{tab:main}
\begin{tabular}{l|c|c|r|r|c|c|r|r}
\toprule
\multicolumn{1}{c|}{\multirow{2}{*}{Metric}}            & \multicolumn{1}{c|}{Consistency(NSI)}                                                & \multicolumn{1}{c|}{Perception(NSI)}                                                    & \multicolumn{1}{c|}{Average} & \multicolumn{1}{c|}{Bitrate} & \multicolumn{1}{c|}{Consistency(AIGI)}                                                & \multicolumn{1}{c|}{Perception(AIGI)} & \multicolumn{1}{c|}{Average}                                                    & \multicolumn{1}{c}{Bitrate} \\ \cline{2-9} 
            & \multicolumn{1}{c|}{(LPIPS$\downarrow$,ClipSIM$\uparrow$)}           & \multicolumn{1}{c|}{(NIQE$\downarrow$,ClipIQA$\uparrow$)}           & \multicolumn{1}{c|}{Avg$\uparrow$} & \multicolumn{1}{c|}{Bpp$\downarrow$}    & \multicolumn{1}{c|}{(LPIPS$\downarrow$,ClipSIM$\uparrow$)}           & \multicolumn{1}{c|}{(NIQE$\downarrow$,ClipIQA$\uparrow$)} & \multicolumn{1}{c|}{Avg$\uparrow$} & \multicolumn{1}{c}{Bpp$\downarrow$}      \\ \midrule
\{T\}JPEG\cite{metric:tra-jpeg}        & (0.5166, 0.7075)                                 & (15.476, 0.2052) &  -1.4036                               & 0.1952                    & (0.5421, 0.8198)                                 & (15.881, 0.2602) & -1.5949                                & 0.1738                   \\
\{T\}WEBP\cite{metric:tra-webp}        & (0.3780, 0.8374)                                 & (7.5893, 0.1514) & 0.8392                                 & 0.1296                    & (0.3578, 0.9160)                                 & (6.9384, 0.1550) & 0.9989                                & 0.1065                   \\
\{T\}VVC\cite{metric:tra-vvc}        & ({\color[HTML]{4472C4} \textbf{0.3577}}, 0.8969)                                 & (6.6469, 0.2390) & 1.7583                                & 0.1035                    & (0.3088, 0.9561)                                 & (6.5056, 0.3173) & 2.0541                                & 0.0951                   \\ \hline
\{N\}BMSHJ\cite{metric:nic-bmshj}       & (0.7543, 0.6406)                                 & (9.0865, 0.2224) & -1.4347                                & 0.0581                    & (0.6859, 0.7739)                                 & (9.1777, 0.2240) & -1.3768                                & 0.0581                   \\
\{N\}MBT\cite{metric:nic-mbt}         & (0.8661, 0.5605)                                 & (10.560, 0.2096) & -2.5149                                & 0.0527                    & (0.7990, 0.6558)                                 & (10.210, 0.2188) & -2.7817                                & 0.0497                   \\
\{N\}CHENG\cite{metric:nic-cheng}       & (0.4625, 0.8198)                                 & (6.4557, 0.2310) & 0.9468                                & 0.0623                    & (0.3723, 0.9316)                                 & (6.7251, 0.2726) & 1.4630                                & 0.0565                   \\ \hline
\{G\}HiFiC\cite{metric:gan-hific}       & (0.3980, {\color[HTML]{FF0000} \textbf{0.9194}}) & (5.6601, 0.4209) & 2.5088                                & 0.0478                    & ({\color[HTML]{FF0000} \textbf{0.2876}}, {\color[HTML]{FF0000} \textbf{0.9663}}) & (5.7907, 0.4205) & {\color[HTML]{4472C4} \textbf{2.6619}}                                & 0.0444                   \\
\{G\}CDC\cite{metric:diff-cdc}         & (0.4692, 0.8564)                                 & (6.2161, 0.2033) & 1.0163                                & 0.0469                    & (0.3416, 0.9414)                                 & (7.3019, 0.2570) & 1.4929                                & 0.0451                   \\
\{G\}PICS\cite{metric:diff-text}        & (0.6080, 0.4668)                                 & ({\color[HTML]{4472C4} \textbf{3.8469}}, {\color[HTML]{FF0000} \textbf{0.6639}}) & 0.9735 & 0.0265                    & (0.6685, 0.7530)                                 & ({\color[HTML]{FF0000} \textbf{4.6612}}, 0.6484) & 0.7417 & 0.0328                   \\ 
\{G\}MISC-1 & (0.5142, 0.8252)                                 & (4.3597, 0.6106) & {\color[HTML]{4472C4} \textbf{2.6162}}                                & 0.0225                    & (0.5026, 0.8921)                                 & (5.0878, {\color[HTML]{4472C4} \textbf{0.7347}}) & 2.4924         & 0.0223                   \\
\{G\}MISC-3   & ({\color[HTML]{FE0000} \textbf{0.3522}}, {\color[HTML]{4472C4} \textbf{0.9106}}) & ({\color[HTML]{FF0000} \textbf{3.8271}}, {\color[HTML]{4472C4}\textbf{0.6612}}) & {\color[HTML]{FF0000} \textbf{3.8957}} & 0.0470                    & {(\color[HTML]{4472C4} \textbf{0.3084}}, {\color[HTML]{4472C4} \textbf{0.9570}}) & ({\color[HTML]{4472C4} \textbf{4.6820}}, {\color[HTML]{FF0000} \textbf{0.7701}}) & {\color[HTML]{FF0000} \textbf{3.8588}} & 0.0446  \\ \bottomrule                
\end{tabular}
\end{table*}

For the existing AIGI database, we use six IQA methods\footnote{BRISQUE \cite{quality:brisque} and NIQE \cite{quality:NIQE} are lower-better values, so these two axes take the reciprocal in Fig. \ref{fig:database}.} to comprehensively evaluate its quality in Fig. \ref{fig:database} including the four indicators in Section \ref{sec:collect}, as well as HyperIQA \cite{quality:HyperIQA} and BRISQUE \cite{quality:brisque}. Higher quality indicates less distortion of the image.
The comparison databases include: DiffusionDB \cite{database:DiffusionDB} and GenImage \cite{database:GenImage}, which contain millions of images generated by different models. They are the largest and most versatile. Pokemon \cite{database:pokemon} is the first AIGI database generated by GAN, which is suitable for early-stage generation task verification. The AGIQA-3K \cite{database:agiqa-3k} and AGIQA-1K \cite{database:agiqa-1k} databases cover three generation architectures: GAN, Auto-Regression, and Diffusion, and are mainly oriented to IQA tasks.
In Fig. \ref{fig:database}, compared to AIGI-SCD, all quality scores of AGIQA-3K/1K lag behind; although other remaining databases occasionally lead AIGI-SGD slightly in some IQA indicators, they all have obvious shortcomings in other indicators. Therefore, these databases from generative models of mixed results are far inferior to AIGI-SCD in overall quality. In all, the characteristics of AIGIs and NSIs are quite different and require targeted compression. Compared with the existing AIGI database, AIGI-SCD has higher quality, less distortion, and \textit{more objective verification of compression algorithms}.

\begin{figure*}[tb]
    \centering
    \subfigure[NSI LPIPS$\downarrow$]{\includegraphics[width = 0.24\textwidth]{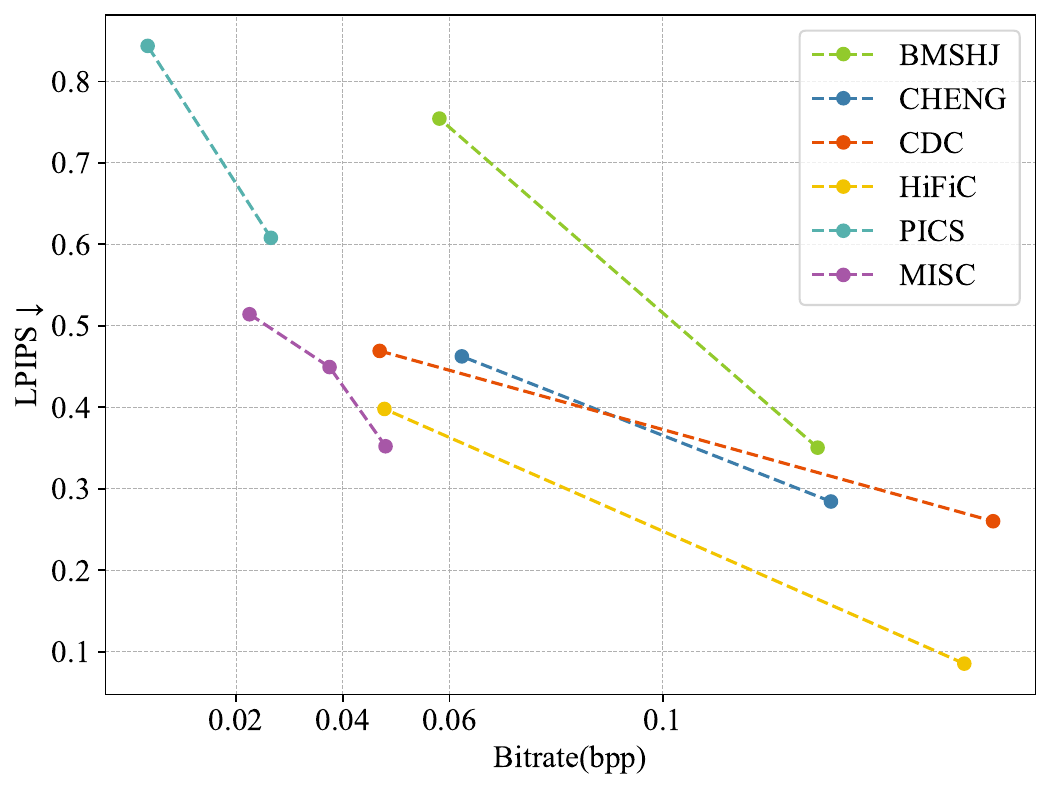}}
    \subfigure[NSI ClipSIM$\uparrow$]{\includegraphics[width = 0.24\textwidth]{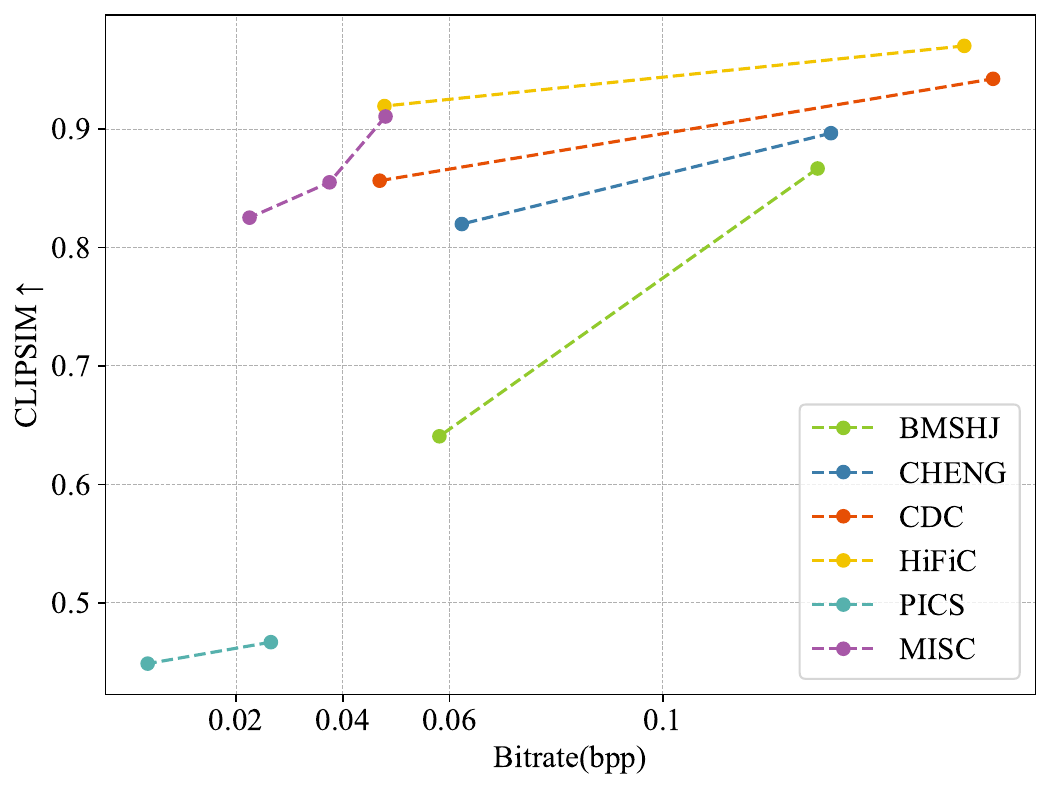}}
    \subfigure[NSI NIQE$\downarrow$]{\includegraphics[width = 0.24\textwidth]{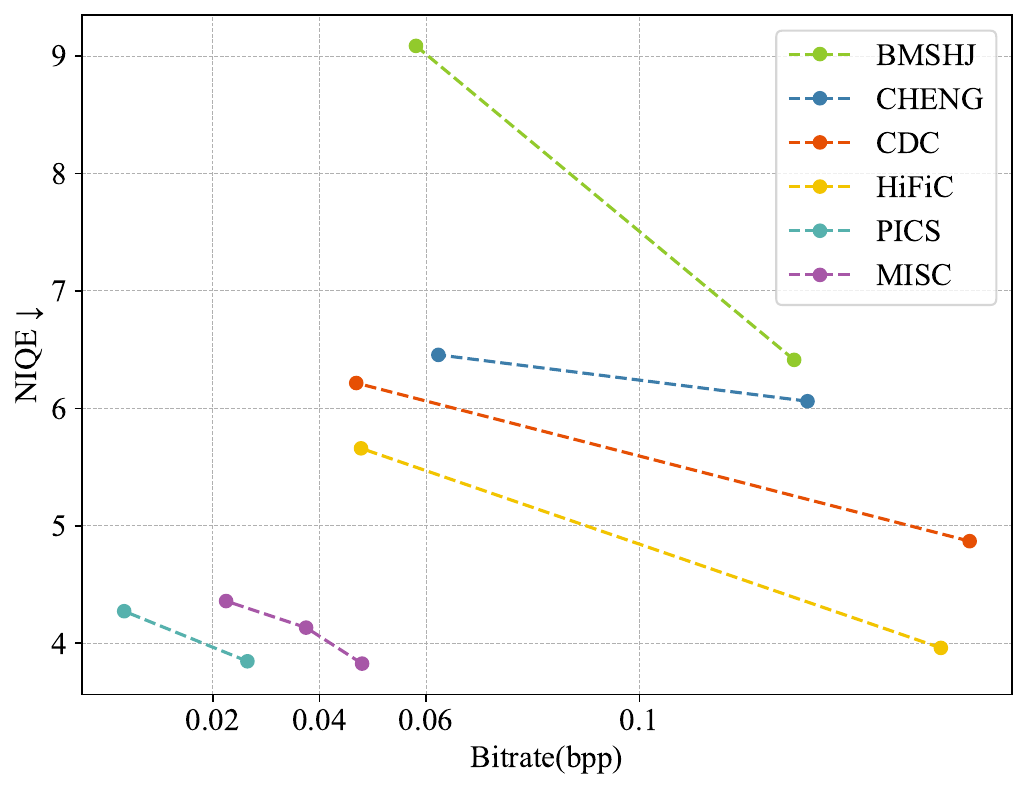}}
    \subfigure[NSI ClipIQA$\uparrow$]{\includegraphics[width = 0.24\textwidth]{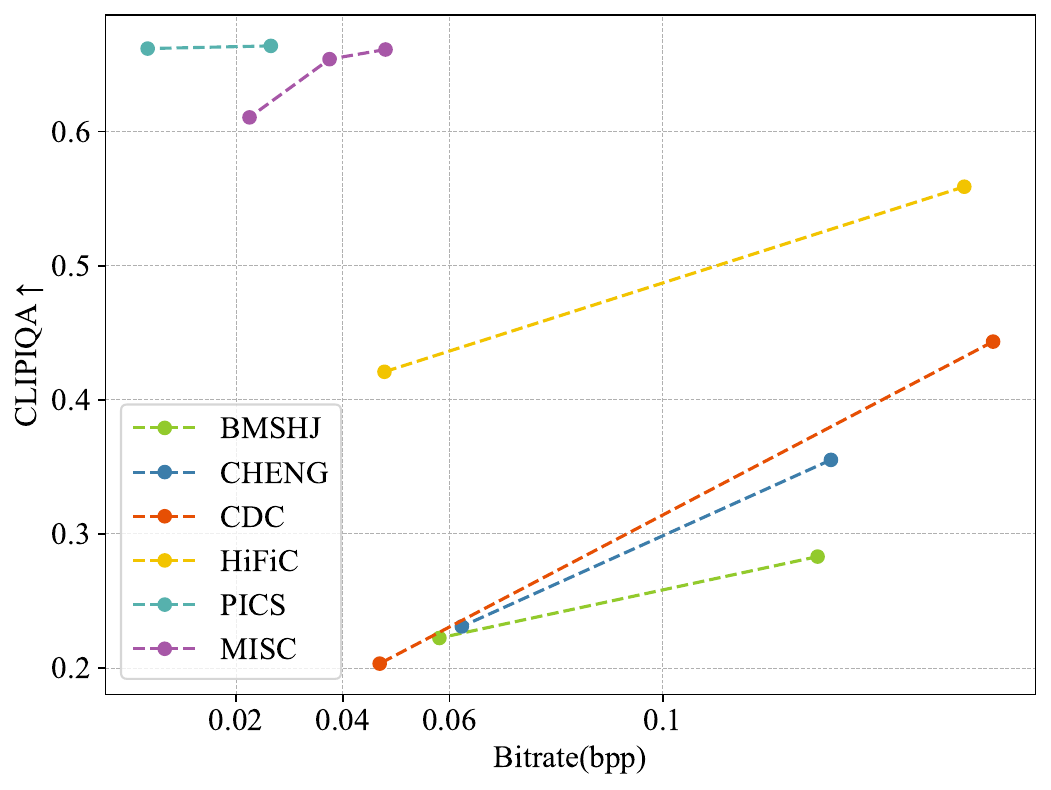}}
    \subfigure[AIGI LPIPS$\downarrow$]{\includegraphics[width = 0.24\textwidth]{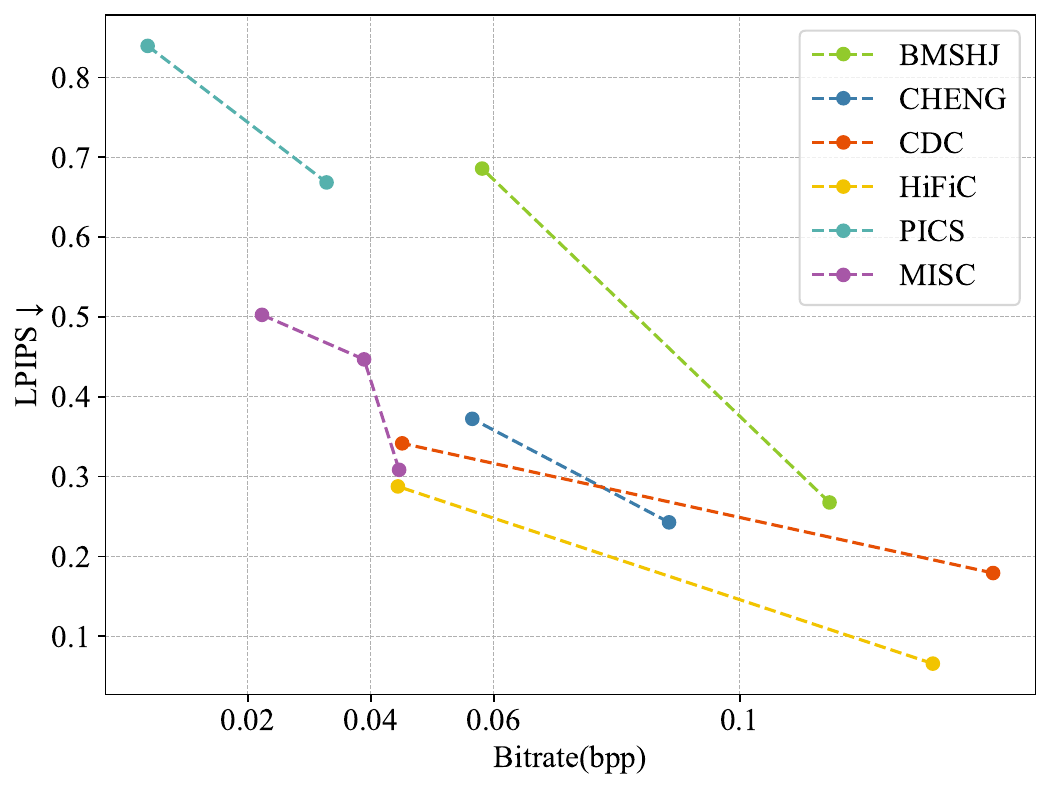}}
    \subfigure[AIGI ClipSIM$\uparrow$]{\includegraphics[width = 0.24\textwidth]{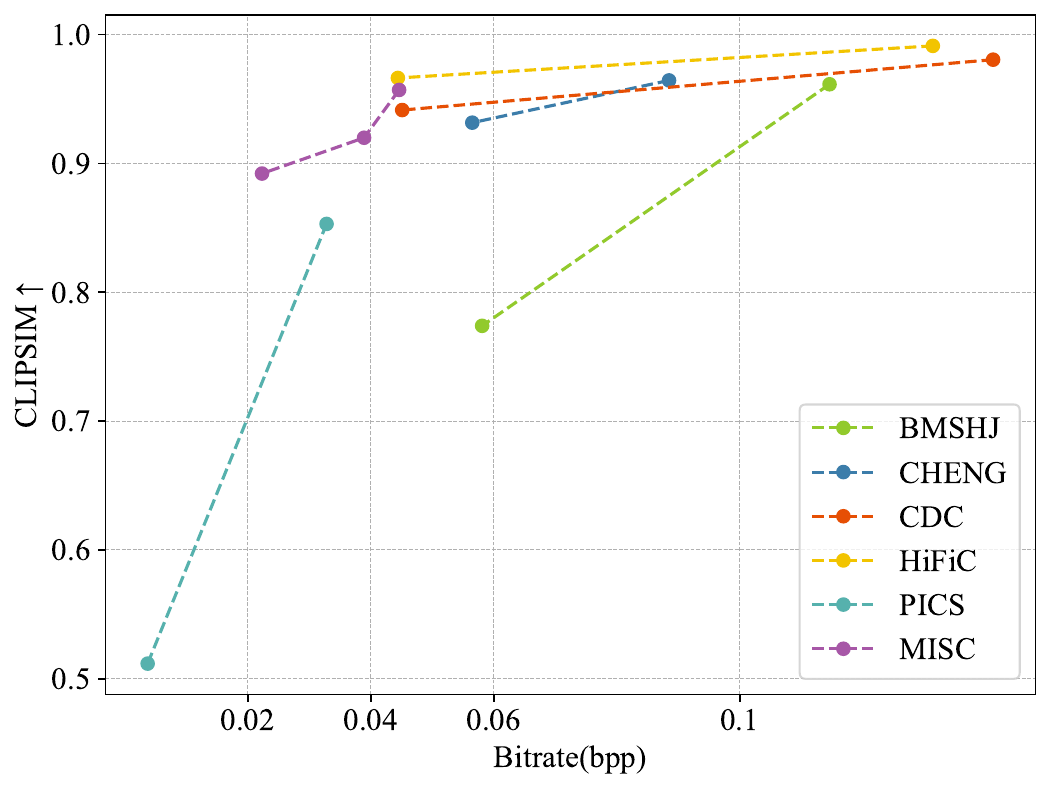}}
    \subfigure[AIGI NIQE$\downarrow$]{\includegraphics[width = 0.24\textwidth]{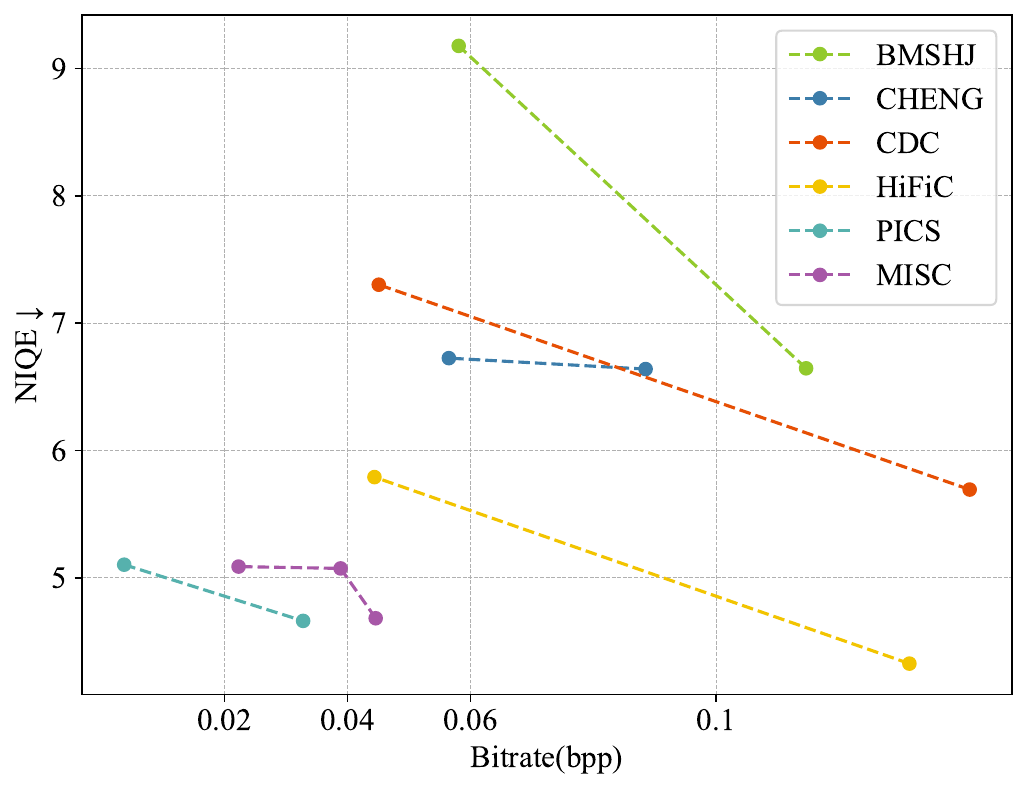}}
    \subfigure[AIGI ClipPIQA$\uparrow$]{\includegraphics[width = 0.24\textwidth]{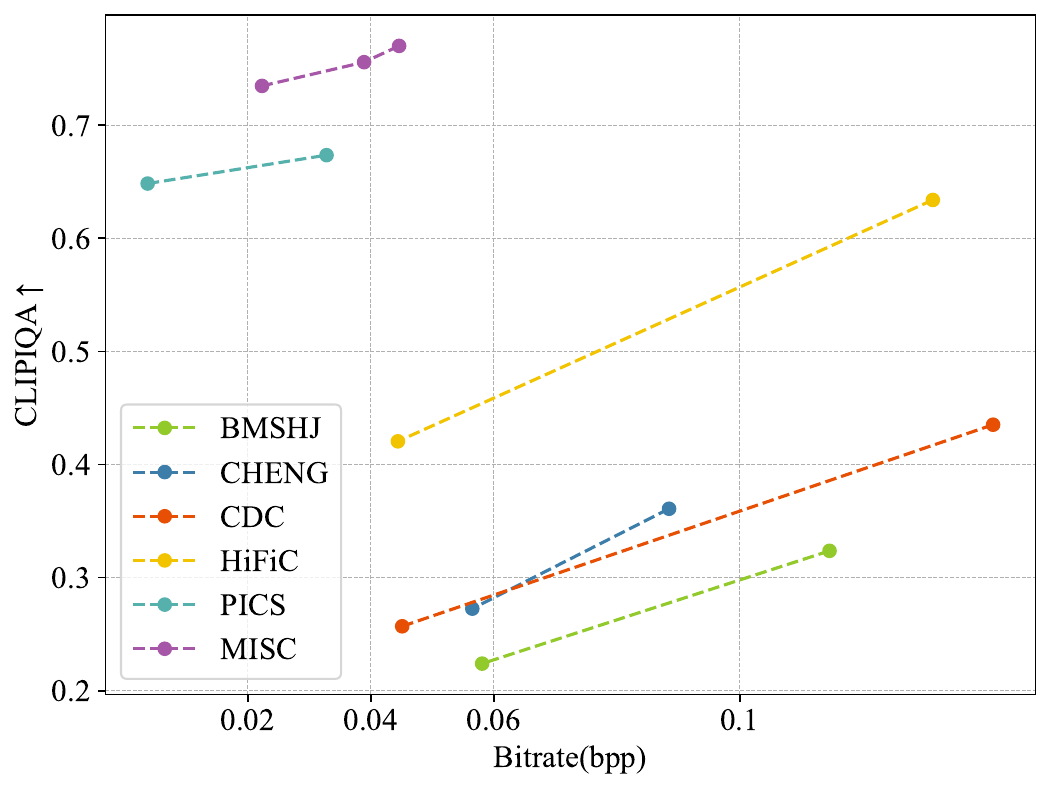}}
    \caption{Variable bitrate of BMSHJ\cite{metric:nic-bmshj}, CHENG\cite{metric:nic-cheng}, CDC\cite{metric:diff-cdc}, HiFiC\cite{metric:gan-hific}, PICS\cite{metric:diff-text}, and MISC we proposed. The left four figures are for consistency while the right four are for perception. Existing methods can only achieve satisfactory results on either consistency or perception. In contrast, MISC provides the possibility for ultra-low bitrate compression at MISC-1 while simultaneously achieving the optimal of all indicators at MISC-3.}
    \label{fig:variable}
\end{figure*}
\begin{table*}[tb]
\caption{Using different compressed content groups in MISC, including three Name-Detail-Map (NDM) groups, a description of all details of the image, and an extremely compressed image bitstream. [Key: {\color[HTML]{FF0000} \textbf{Best}}; {\color[HTML]{4472C4}\textbf{Selected Content} \faCheckSquare}].}
\label{tab:ablation}
\centering
\subtable[MISC validation on CLIC2020 \cite{database:clic} database (Left, Right): (MISC-1, MISC-2)]{\begin{tabular}{lcc|cc|cc|c}
\toprule
\multicolumn{3}{c|}{Content}   & \multicolumn{2}{c|}{Consistency}   & \multicolumn{2}{c|}{Perception}       & Bitrate         \\ \hline
\multicolumn{1}{c}{NDM} & Detail all & Bitstream & LPIPS$\downarrow$           & ClipSIM$\uparrow$         & NIQE$\downarrow$            & ClipIQA$\uparrow$         & Bpp$\downarrow$             \\ \midrule
{\color[HTML]{4472C4} \faCheckSquare}{\color[HTML]{4472C4} \faCheckSquare}{\color[HTML]{4472C4} \faCheckSquare} & {\color[HTML]{4472C4} \faCheckSquare}        & {\color[HTML]{4472C4} \faCheckSquare}       & ({\color[HTML]{FF0000} \textbf{0.5142}}, {\color[HTML]{FF0000} \textbf{0.4494}}) & ({\color[HTML]{FF0000} \textbf{0.8252}}, {\color[HTML]{FF0000} \textbf{0.8550}}) & (4.3597, 4.1336) & (0.6106, {\color[HTML]{FF0000} \textbf{0.6540}}) & (0.0225, 0.0375) \\
{\color[HTML]{4472C4} \faCheckSquare}{\color[HTML]{4472C4} \faCheckSquare}{\color[HTML]{4472C4} \faCheckSquare} &            & {\color[HTML]{4472C4} \faCheckSquare}       & (0.6244, 0.5648) & (0.7764, 0.8184) & (8.8310, 8.0048) & (0.2587, 0.3031) & (0.0187, 0.0337) \\
{\color[HTML]{4472C4} \faCheckSquare}{\color[HTML]{4472C4} \faCheckSquare} &            & {\color[HTML]{4472C4} \faCheckSquare}       & (0.6311, 0.5733) & (0.7725, 0.8149) & (8.6842, 7.9538) & (0.2372, 0.2753) & (0.0180, 0.0330) \\
{\color[HTML]{4472C4} \faCheckSquare} &            & {\color[HTML]{4472C4} \faCheckSquare}       & (0.6387, 0.5856) & (0.7764, 0.8125) & (8.8231, 8.0993) & (0.2114, 0.2335) & (0.0172, 0.0323) \\
    & {\color[HTML]{4472C4} \faCheckSquare}        & {\color[HTML]{4472C4} \faCheckSquare}       & (0.5222, 0.4690) & (0.8149, 0.8530) & ({\color[HTML]{FF0000} \textbf{4.0333}}, {\color[HTML]{FF0000} \textbf{3.8690}}) & ({\color[HTML]{FF0000} \textbf{0.6335}}, 0.6368) & (0.0202, 0.0353) \\
{\color[HTML]{4472C4} \faCheckSquare}{\color[HTML]{4472C4} \faCheckSquare}{\color[HTML]{4472C4} \faCheckSquare} &  {\color[HTML]{4472C4} \faCheckSquare}      &           & 0.7760          & 0.7613          & 4.5878          & 0.6315          & 0.0061  \\ \bottomrule       
\end{tabular}}
\subtable[MISC validation on AIGI-SCD database (Left, Right): (MISC-1, MISC-2)]{\begin{tabular}{lcc|cc|cc|c}
\toprule
\multicolumn{3}{c|}{Content}   & \multicolumn{2}{c|}{Consistency}   & \multicolumn{2}{c|}{Perception}       & Bitrate         \\ \hline
\multicolumn{1}{c}{NDM} & Detail all & Bitstream & LPIPS$\downarrow$           & ClipSIM$\uparrow$         & NIQE$\downarrow$            & ClipIQA$\uparrow$         & Bpp$\downarrow$             \\ \midrule
{\color[HTML]{4472C4} \faCheckSquare}{\color[HTML]{4472C4} \faCheckSquare}{\color[HTML]{4472C4} \faCheckSquare} & {\color[HTML]{4472C4} \faCheckSquare}        & {\color[HTML]{4472C4} \faCheckSquare}       & ({\color[HTML]{FF0000} \textbf{0.5026}}, {\color[HTML]{FF0000} \textbf{0.4468}}) & ({\color[HTML]{FF0000} \textbf{0.8921}}, {\color[HTML]{FF0000} \textbf{0.9204}}) & (5.0878, 5.0734) & ({\color[HTML]{FF0000} \textbf{0.7347}}, {\color[HTML]{FF0000} \textbf{0.7557}}) & (0.0223, 0.0389) \\
{\color[HTML]{4472C4} \faCheckSquare}{\color[HTML]{4472C4} \faCheckSquare}{\color[HTML]{4472C4} \faCheckSquare} &            & {\color[HTML]{4472C4} \faCheckSquare}       & (0.5299, 0.4651) & (0.8643, 0.9136) & (12.401, 11.784) & (0.3300, 0.4233) & (0.0188, 0.0354) \\
{\color[HTML]{4472C4} \faCheckSquare}{\color[HTML]{4472C4} \faCheckSquare} &            & {\color[HTML]{4472C4} \faCheckSquare}       & (0.5390, 0.4769) & (0.8638, 0.9082) & (12.780, 11.928) & (0.3051, 0.3705) & (0.0180, 0.0347) \\
{\color[HTML]{4472C4} \faCheckSquare} &            & {\color[HTML]{4472C4} \faCheckSquare}       & (0.5469, 0.4862) & (0.8613, 0.9023) & (13.136, 12.391) & (0.2902, 0.3374) & (0.0173, 0.0340) \\
    & {\color[HTML]{4472C4} \faCheckSquare}        & {\color[HTML]{4472C4} \faCheckSquare}       & (0.5480, 0.4671) & (0.8794, 0.9199) & (4.6725, 4.3572) & (0.7136, 0.7251) & (0.0200, 0.0367) \\
{\color[HTML]{4472C4} \faCheckSquare}{\color[HTML]{4472C4} \faCheckSquare}{\color[HTML]{4472C4} \faCheckSquare} &  {\color[HTML]{4472C4} \faCheckSquare}      &           & 0.7416          & 0.7656          & {\color[HTML]{FF0000} \textbf{4.2413}}          & 0.6901          & 0.0058  \\ \bottomrule       
\end{tabular}}
\end{table*}
\section{Expriment}
\subsection{Experiment Settings}
To assess the efficacy of the proposed MISC method across diverse image types, we conducted performance evaluations on two distinct databases: the commonly used CLIC2020\footnote{As the maximum output of current GIC metrics is 1,024 pixels, we $2\times$ downsampled images larger than this bound.} database for NSIs compression \cite{database:clic}, and the AIGI-SCD database specifically designed for AIGIs. Following the standard partitioning, the training/test image distribution comprises 585/41 images for CLIC2020 and 450/50 images for AIGI-SCD.
In our methodology, we kept the LMM and map encoders, as well as the T2I component of the decoder (utilizing default parameters of GPT-4 Vision \cite{intro:gpt4}, CLIP \cite{model:clip}, and DiffBIR \cite{model:diffbir}) frozen, and focused on fine-tuning the low bitrate image encoder/decoder. While the encoder/decoder architecture aligns with existing NIC and GIC frameworks, the parameters underwent an intensive ultra-low fine-tuning process. Specifically, this process initiated the model at the lowest bitrate mode, followed by a tenfold increase in the bitrate weight of the loss function ($\lambda$ reduced to $\frac{1}{10}$ of its original value), enabling training for extreme compression with a learning rate of $10^{-4}$.
MISC has three compression levels for dynamic adjustment. The first two levels (MISC-1/2) are for $0.02\sim0.03$ bpp, while we activate all encoders; the last level (MISC-3) is for $0.04\sim0.05$ bpp. This relatively higher bitrate can accommodate more image details, and using items to guide diffusion is not as important as before. To save bitrate, we discarded the information of these three NDM groups, by deactivating the item and detail questions in the LMM encoder, and the whole map encoder, then only used other modules for compression.
In our comparative analysis, MISC is benchmarked against nine mainstream low-bitrate image compression methods, including traditional JPEG \cite{metric:tra-jpeg}, WEBP \cite{metric:tra-webp}, VVC (Intra frame mode) \cite{metric:tra-vvc}, NIC's BMSHJ \cite{metric:nic-bmshj}, MBT \cite{metric:nic-mbt}, CHENG \cite{metric:nic-cheng}, and GIC's HiFiC \cite{metric:gan-hific}, CDC \cite{metric:diff-cdc}, PICS \cite{metric:diff-text}. As ultra-low bitrates are not their best scenario (excluding PICS), we applied the ultra-low fine-tuning approach above to all trainable models for a fair comparison. The experiments were conducted using NVIDIA RTX 4090 GPUs with the Adam optimizer \cite{other:adam}.

In our assessment of compression performance, we employ a comprehensive set of metrics to address both consistency and perception requirements. As discussed in Section \ref{sec:relate-iqa}, semantic indicators become crucial at ultra-low bitrates for characterizing differences between the compressed image and the ground truth, surpassing the significance of pixel-level metrics such as PSNR and SSIM. Therefore, we utilize LPIPS \cite{iqa:lpips}, a widely used visual metric based on the Human Visual System (HVS), to quantify the distortion post-compression. Additionally, following prior work in LMM semantic compression \cite{metric:diff-text}, we utilize ClipSIM \cite{model:clip} to calculate the cosine distance of CLIP embeddings, assessing the similarity of semantic features between images.
For perception evaluation, we utilize the IQA method NIQE \cite{quality:NIQE} to evaluate low-level distortions like blur and noise. Furthermore, we incorporate the emerging IAA method ClipIQA \cite{quality:CLIPIQA} to measure human aesthetic satisfaction with the image. To provide a comprehensive representation of performance across all metrics, we calculate a normalized average of the four indicators mentioned above using equations (\ref{equ:avg1}) and (\ref{equ:avg2}) to offer an overall assessment of both consistency and perception aspects.

\begin{figure*}[tb]
    \centering
    \subfigure[Ground Truth]{\includegraphics[width = 0.16\textwidth]{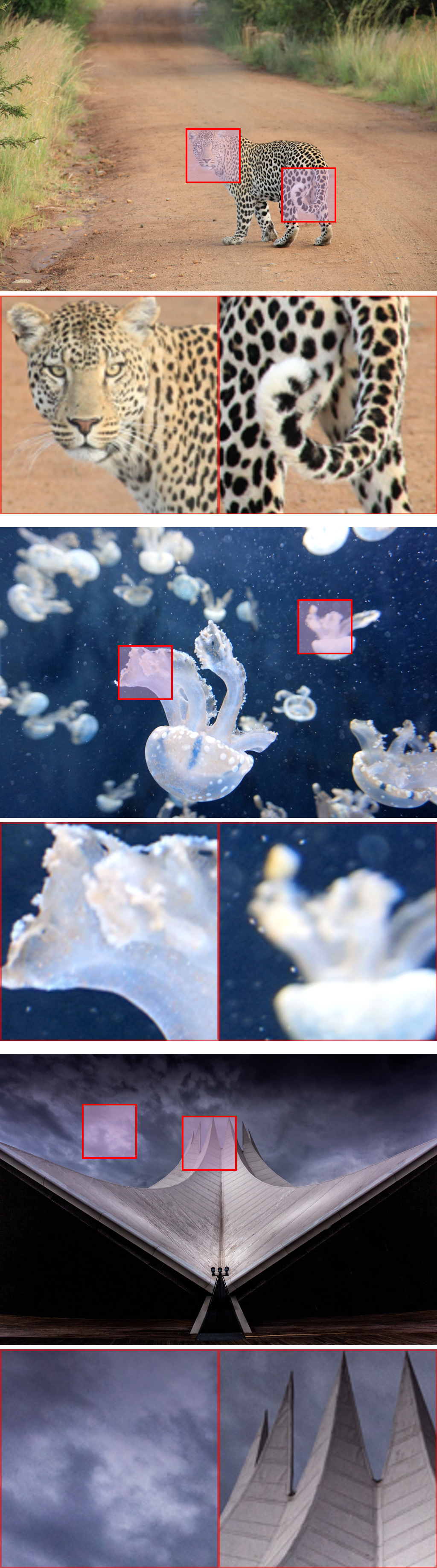}}
    \subfigure[CHENG(0.0623 bpp)]{\includegraphics[width = 0.16\textwidth]{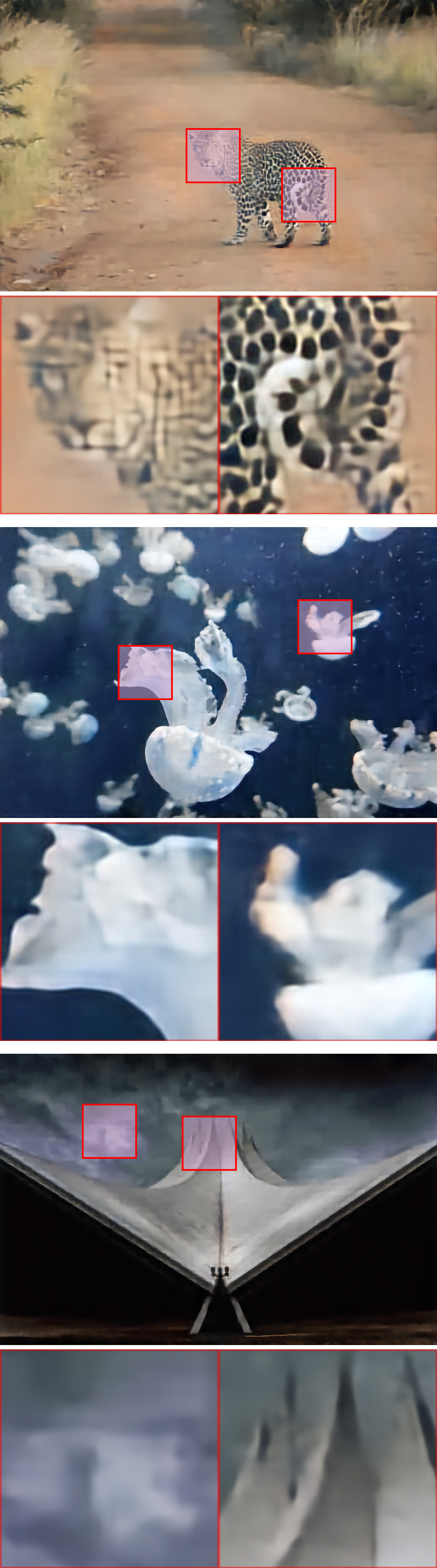}}
    \subfigure[PICS(0.0265 bpp)]{\includegraphics[width = 0.16\textwidth]{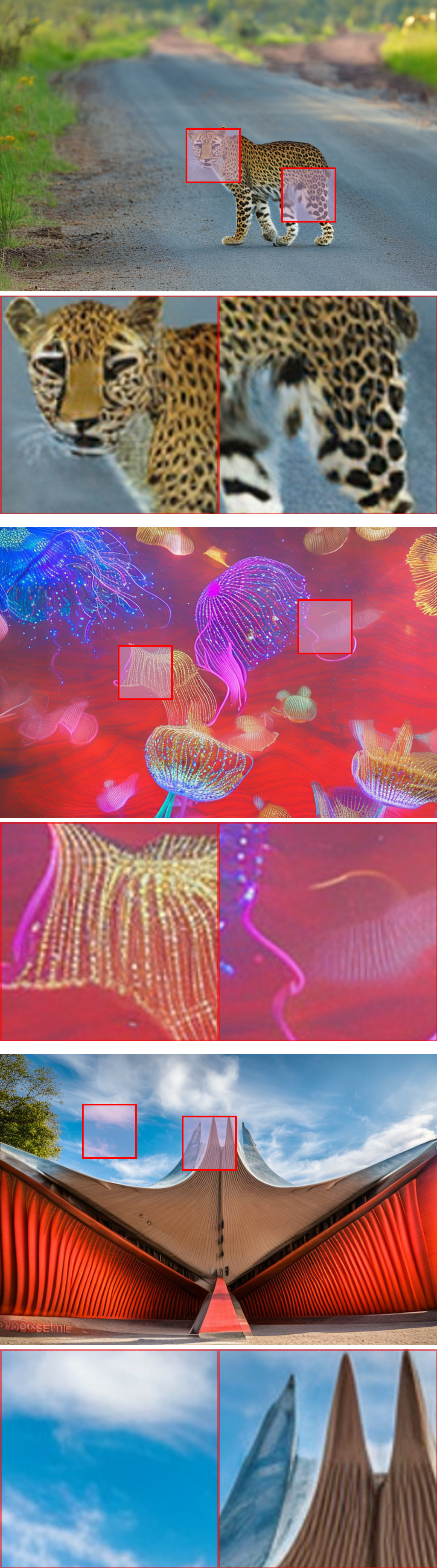}}
    \subfigure[MISC-1(0.0225 bpp)]{\includegraphics[width = 0.16\textwidth]{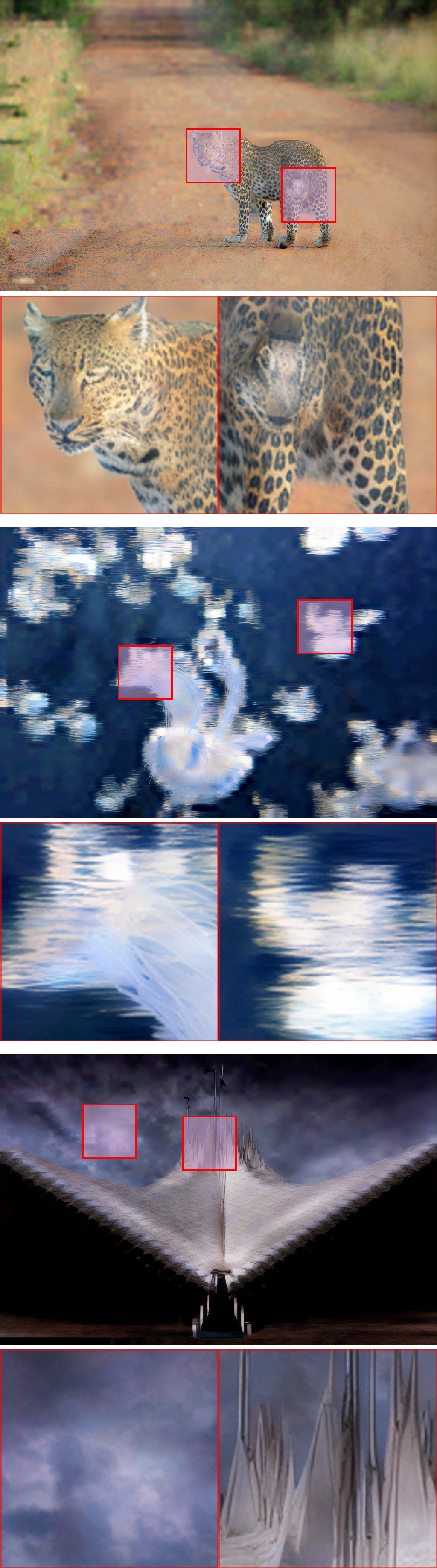}}
    \subfigure[MISC-2(0.0375 bpp)]{\includegraphics[width = 0.16\textwidth]{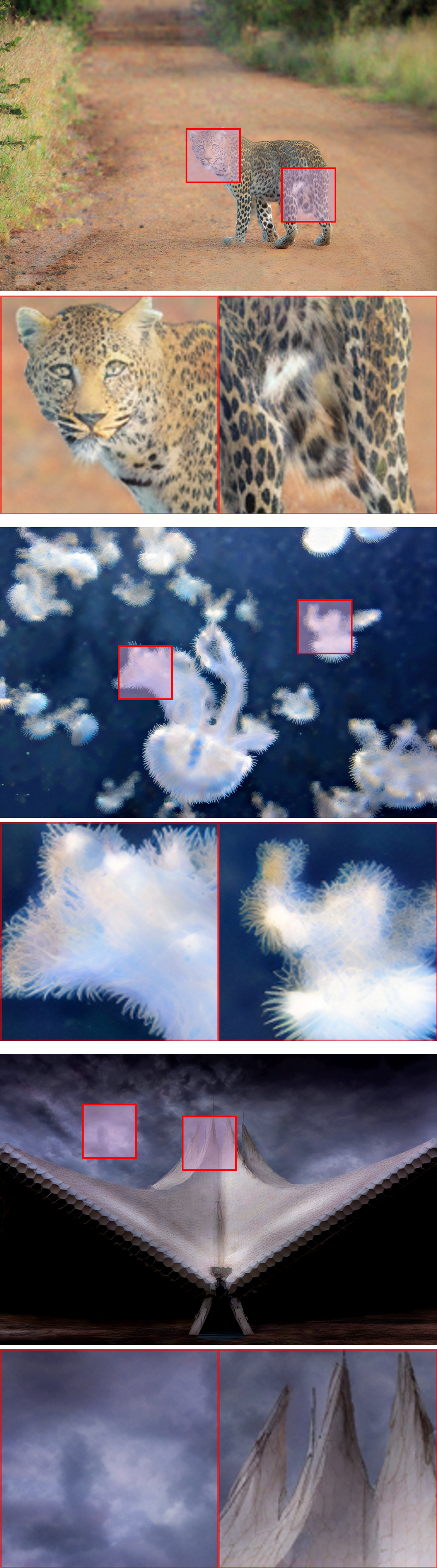}}
    \subfigure[MISC-3(0.0470 bpp)]{\includegraphics[width = 0.16\textwidth]{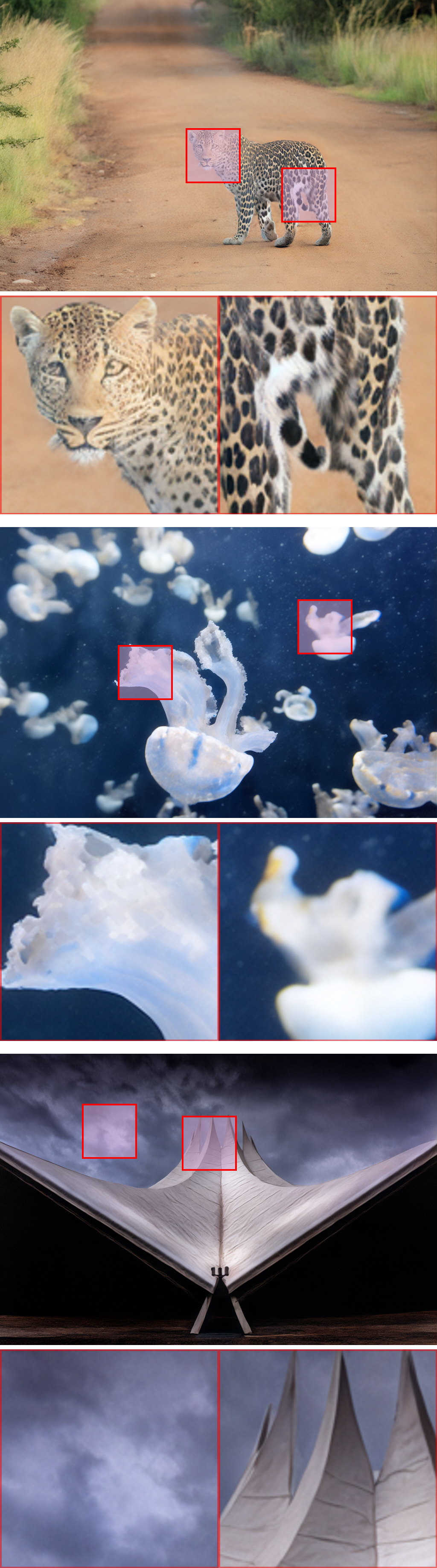}}
    \caption{Qualitative experiment of NSI compression. CHENG \cite{metric:nic-cheng} is suitable for only consistency, PICS \cite{metric:diff-text} experts in only perception, and MISC performs the best in both indicators.}
    \label{fig:vis-nsi}
\end{figure*}

\begin{figure*}[tb]
    \centering
    \subfigure[Ground Truth]{\includegraphics[width = 0.16\textwidth]{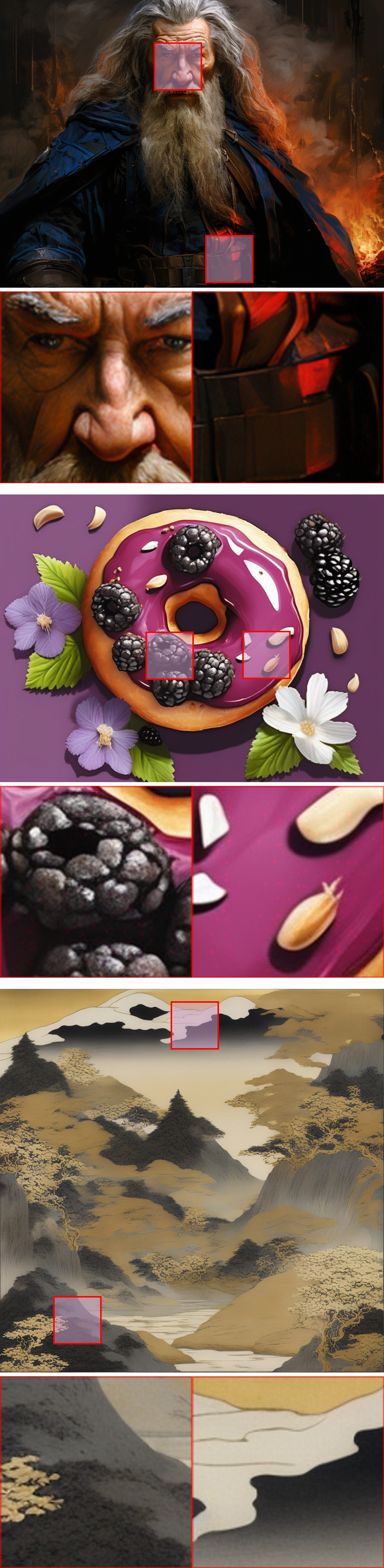}}
    \subfigure[CHENG(0.0565 bpp)]{\includegraphics[width = 0.16\textwidth]{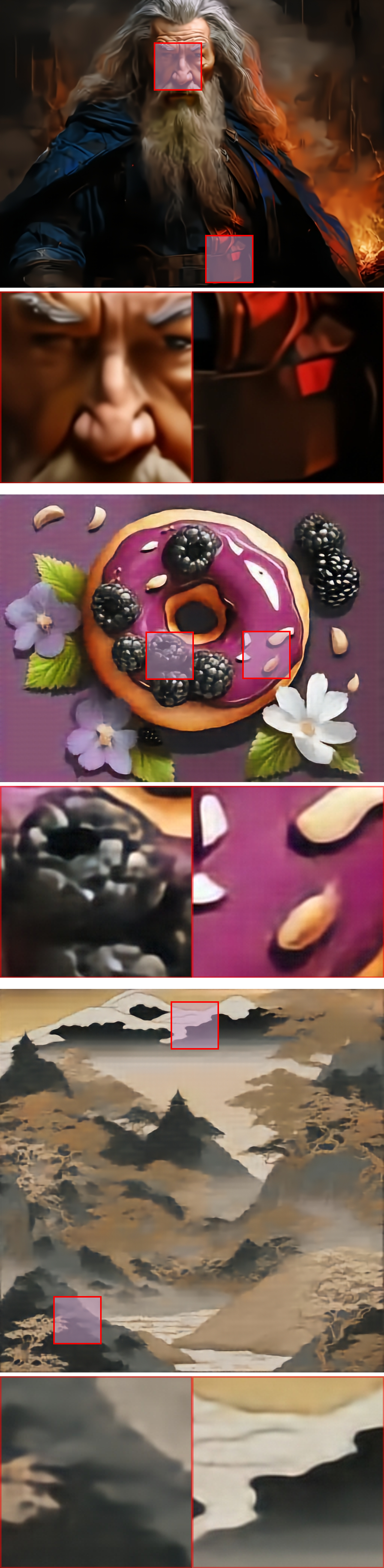}}
    \subfigure[PICS(0.0328 bpp)]{\includegraphics[width = 0.16\textwidth]{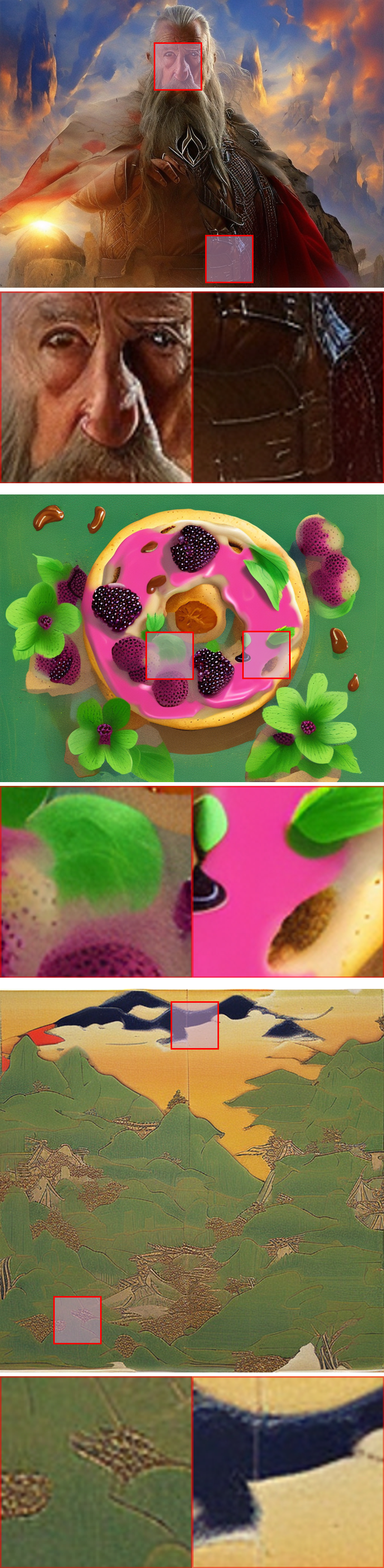}}
    \subfigure[MISC-1(0.0223 bpp)]{\includegraphics[width = 0.16\textwidth]{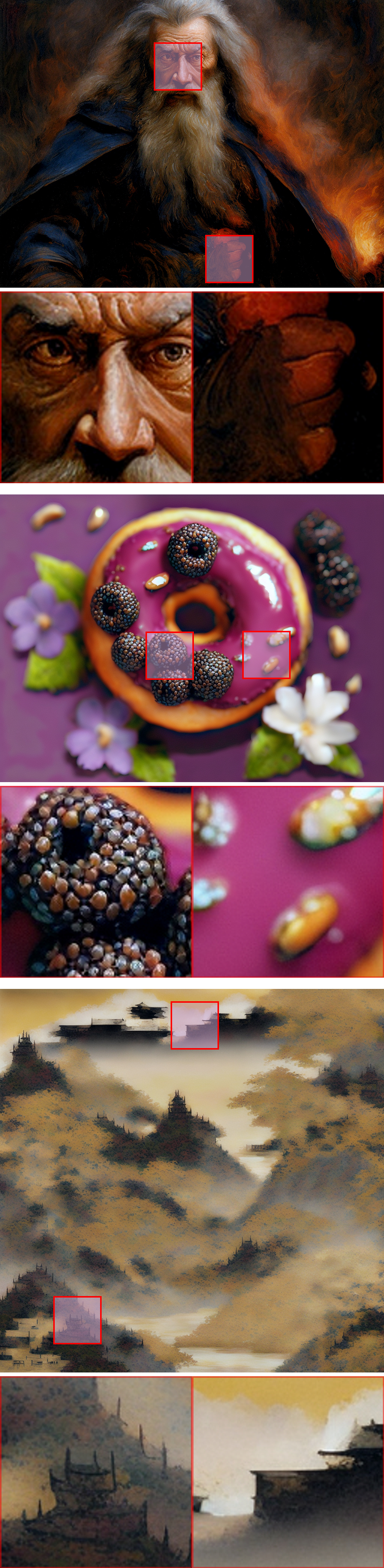}}
    \subfigure[MISC-2(0.0389 bpp)]{\includegraphics[width = 0.16\textwidth]{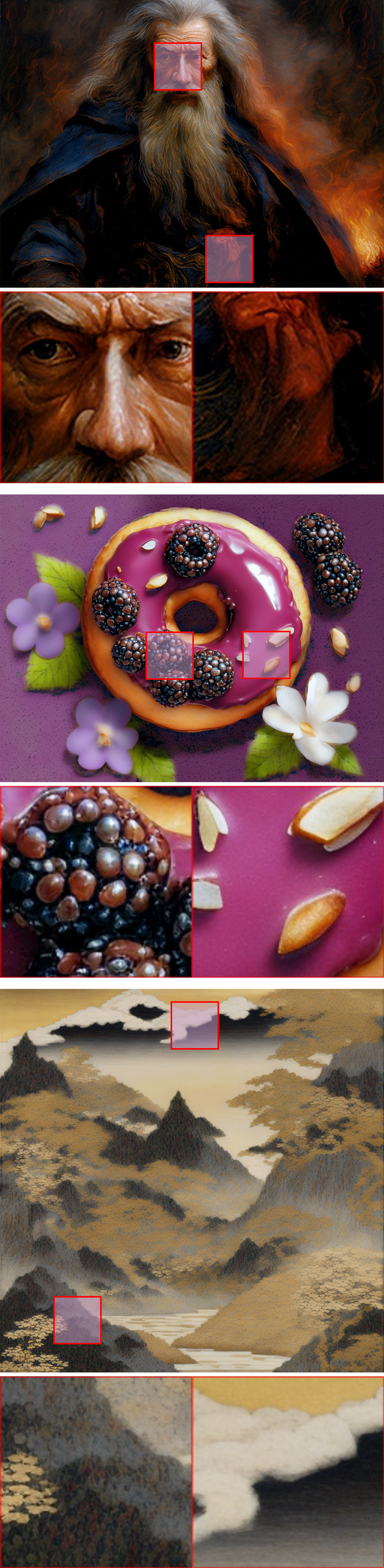}}
    \subfigure[MISC-3(0.0446 bpp)]{\includegraphics[width = 0.16\textwidth]{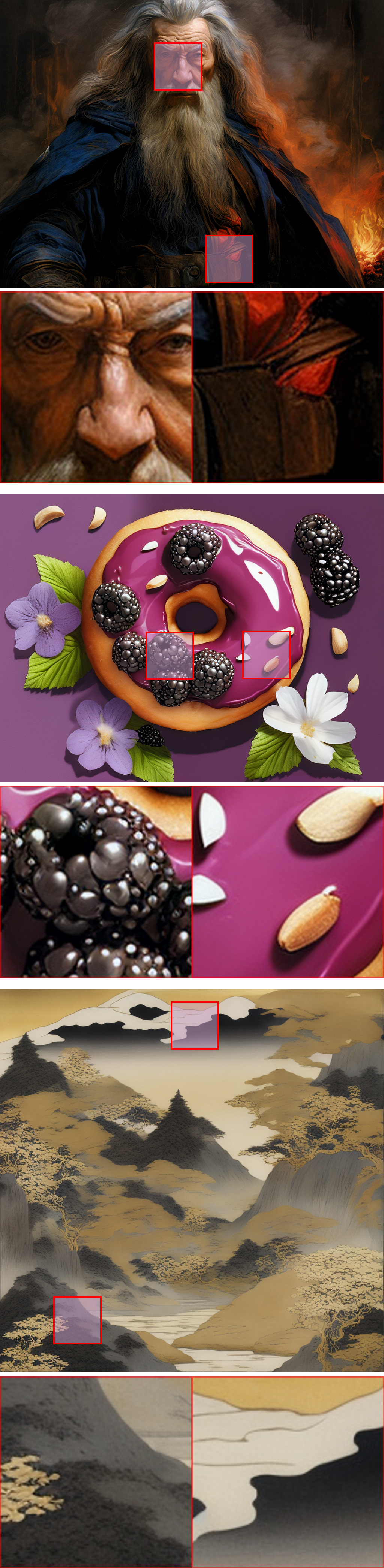}}
    \caption{Qualitative experiment of AIGI compression. CHENG \cite{metric:nic-cheng} is suitable for only consistency, PICS \cite{metric:diff-text} experts in only perception, and MISC performs the best in both indicators.}
    \label{fig:vis-aigi}
\end{figure*}

\subsection{Experiment Results and Discussion}
At ultra-low/low bitrate, we compared other advanced methods against MISC from the lowest (MISC-1) to the highest (MISC-3) bitrate levels respectively. As depicted in TABLE \ref{tab:main}, all the consistency and perception indicators of MISC-3 rank as the best or second best. Considering both consistency and perceptual quality collectively, MISC-3 achieves a normalized average score exceeding 3.8, significantly outperforming other methods. Even for MISC-1, it matches HiFiC in consistency while utilizing only 50\% of its bitrate overhead.
Examining each indicator independently, methods like VVC, HiFiC, and CDC can achieve high consistency with the original image but exhibit poor perceptual quality. For low-level vision, NIQE hovers around 6, while ClipIQA struggles to surpass 0.4 in aesthetics. In contrast, PICS offers higher perceptual quality but sacrifices consistency with the original image, ranking even the lowest in LPIPS and ClipIQA. Here, our MISC-3 matches HiFiC in consistency and surpasses NIQE/ClipIQA by approximately 1.8/0.24 in perception. Furthermore, it equals to PICS in perception and surpasses LPIPS/ClipSIM by around 0.35/0.45.
In addition to its superior performance around 0.05 bpp, our approach excels in ultra-low bitrate compression below 0.024 bpp. Traditional compression methods have inherent bitrate limitations, and learning-based NIC/GIC methods struggle to train and converge at such low bitrates.
In contrast, MISC-1 achieves acceptable consistency results for the first time, with slightly inferior LPIPS/ClipSIM compared to CHENG but nearly three times bitrate savings, while maintaining superior image perceptual quality.
In summary, MISC has achieved a balance for the first time in addressing the conflicting optimization goals of extreme compression at low/ultra-low bitrates while reconstructing images with limited information, enhancing high-quality details consistent with the original image, and resolving the consistency-perception dilemma.

Fig. \ref{fig:variable} further validates the performance of MISC and other methods across different bitrates. MISC not only leads at the 0.05 bpp bitrate but also maintains a significant advantage in perception compared to methods in the $0.1\sim0.2$ bpp range, with only a 30\% bitrate overhead. Fig. \ref{fig:vis-nsi} and \ref{fig:vis-aigi} showcase visual compression examples of MISC on NSIs and AIGIs, encompassing characters, items, and backgrounds. Compared to traditional methods, MISC generates rich high-quality details and aligns more closely with the ground truth. Internally comparing different levels of MISC, MISC-1 produces outlines roughly consistent with the ground truth but exhibits some freedom in character expressions, object colors, and building shapes. MISC-2 improves on consistency with the ground truth but still lacks accuracy in rendering details like clothing, sky, and mountains. MISC-3 achieves near-complete consistency with the ground truth, with minimal texture differences upon zooming in that do not impact visual quality. In essence, MISC offers dynamic bitrate adjustment capabilities and demonstrates versatility in low/ultra-low bitrate scenarios.

Besides, regarding compressed image content, TABLE \ref{tab:main} and Fig. \ref{fig:variable} illustrate that AIGI compression achieves higher consistency than NSI at equivalent bitrates, along with superior perceptual quality in aesthetics. However, existing compression algorithms fall short in low-level quality. Various methods on AIGI only reach 80\% of NSI in terms of NIQE, attributed to both inherent low-level defects and the limited understanding of compression algorithms on AIGI. Notably, despite the low-level quality of NIQE, MISC excels in aesthetic quality, achieving a ClipIQA score above 0.7 even at the lowest MISC-1. In conclusion, beyond NSI, providing viewers with high perceptual quality and consistency for AIGI poses a significant challenge for future compression metrics.

\subsection{Ablation Study}

To validate the contributions of the different compressed content in MISC, an ablation study was conducted, and the results are presented in Table TABLE \ref{tab:ablation}. The factors are specified as ($\romannumeral 1$) Three  Name-Detail-Map (NDM) group for items. ($\romannumeral 2$) A text description of all details in the image. ($\romannumeral 3$) A bitstream characterizing the extremely compressed image content. Considering that MISC-3 does not depend on NDM groups, we use MISC-1 and MISC-2, discard the above three contents during the compression process, and evaluate the results in terms of consistency and perception. The results show that [\textbf{Detail all}] is the most important content for both NSI and AIGI. Without its guidance, both consistency and perception performance decrease across the board.
In addition, [\textbf{Bitstream}] also lays the foundation of MISC. Although its absence will not affect the Perception indicator, it will lead to a significant drop in Consistency. 
The only contentious issue lies in the utilization of [\textbf{NDM}] content. For perception, using NDM enhances aesthetic quality but diminishes low-level quality. Conversely, for consistency, omitting NDM results in an overall performance decline. As Fig. \ref{fig:variable} illustrates MISC already has better perception beyond consistency. Given that incorporating three NDM groups only necessitates a bitrate overhead of 0.002 bpp, adding NDM groups can enhance consistency with minimal sacrifice in perception. In conclusion, eliminating any single content leads to performance deterioration, affirming their collective contribution to the final compression performance.

\subsection{User Study}
\begin{figure}[tb]
    \centering
    \includegraphics[width = 0.48\textwidth]{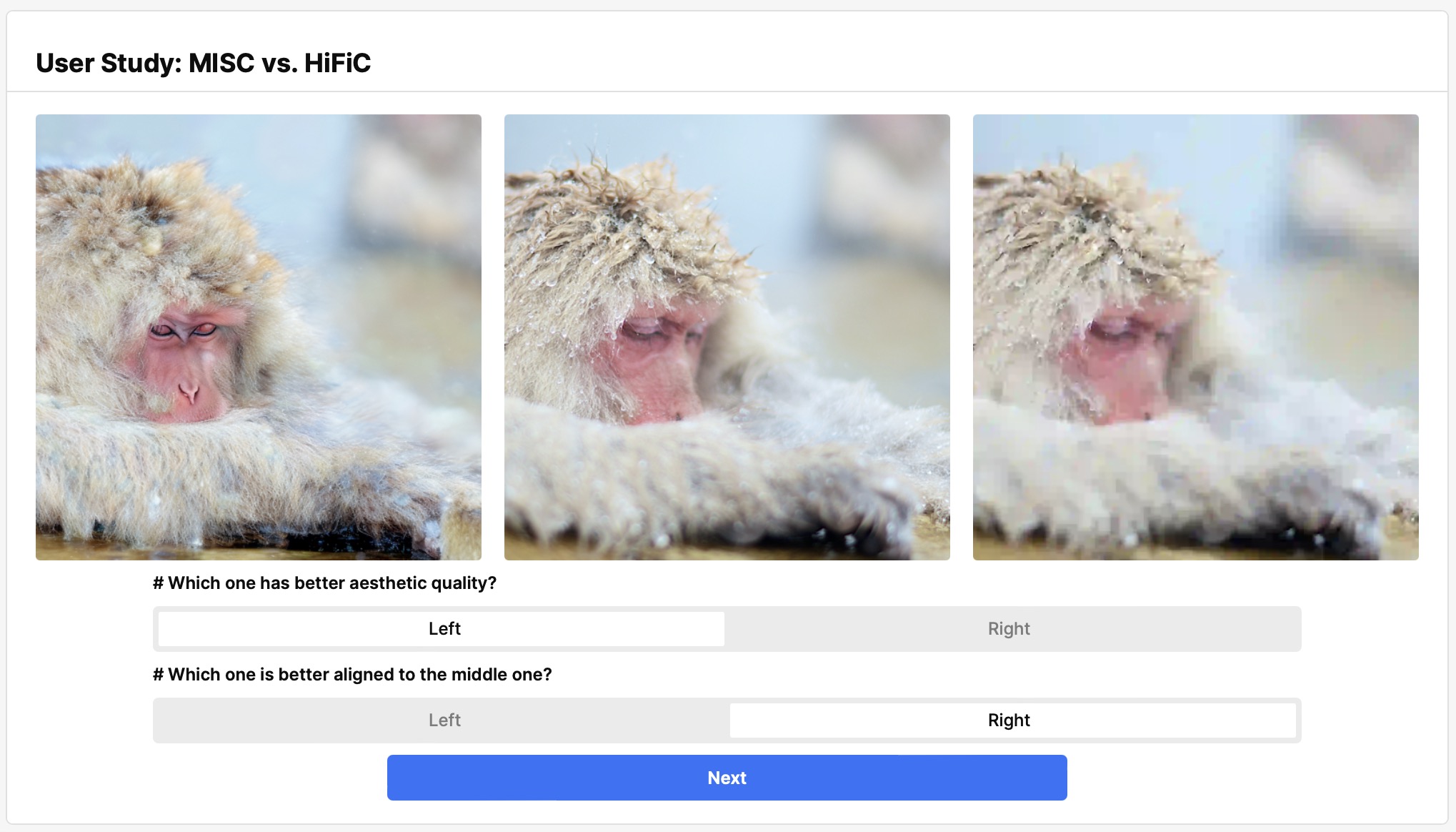}
    \caption{User interface for choosing preference in terms of consistency/perception. The ground truth image in the middle is compressed by different metrics on the left/right.}
    \label{fig:html}
\end{figure}

\begin{figure}[tb]
    \centering
    \subfigure[NSI Consistency]{\includegraphics[width = 0.24\textwidth]{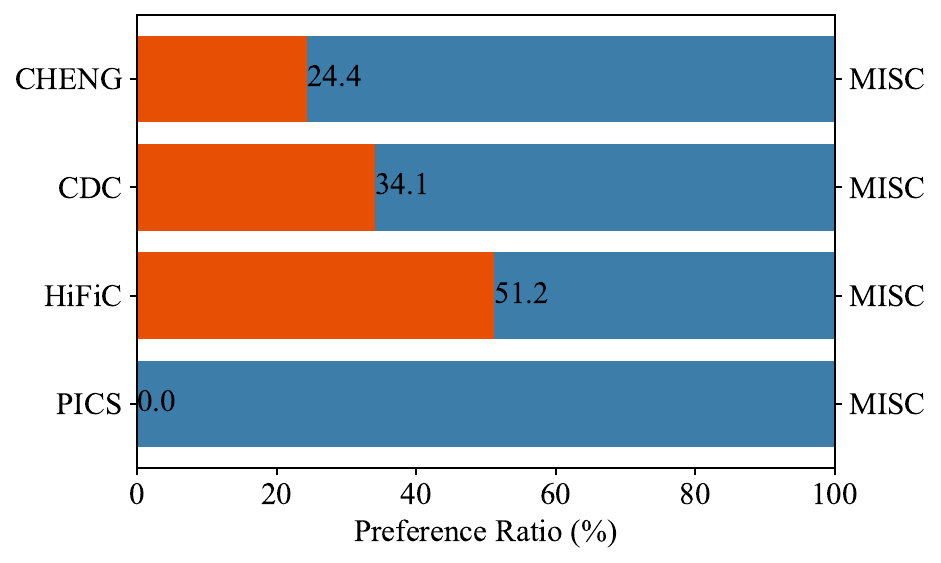}}
    \subfigure[NSI Perception]{\includegraphics[width = 0.24\textwidth]{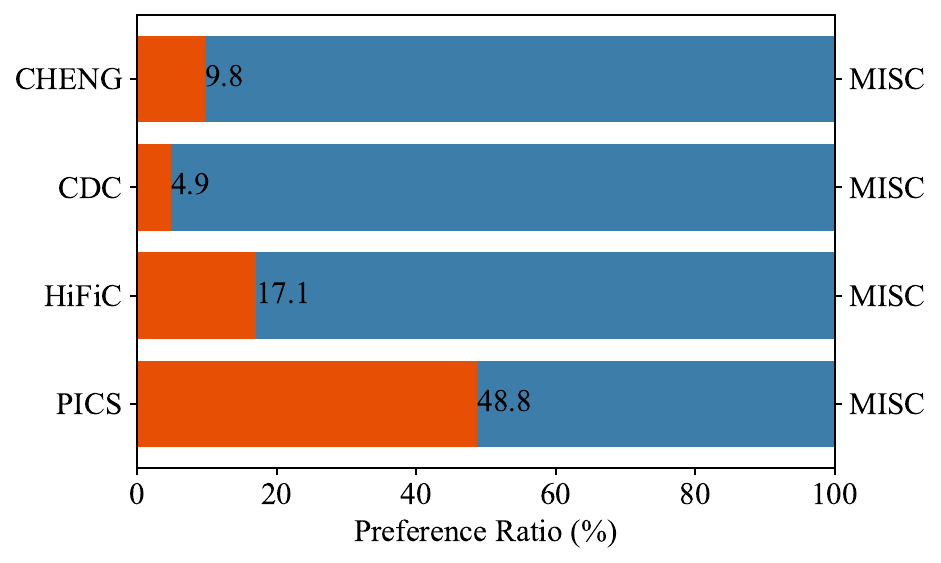}}
    \subfigure[AIGI Consistency]{\includegraphics[width = 0.24\textwidth]{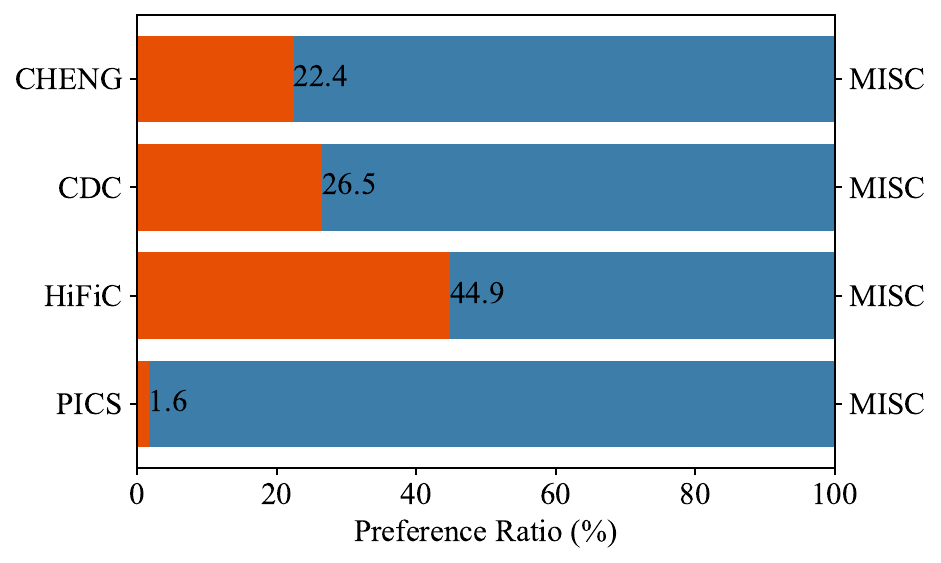}}
    \subfigure[AIGI Perception]{\includegraphics[width = 0.24\textwidth]{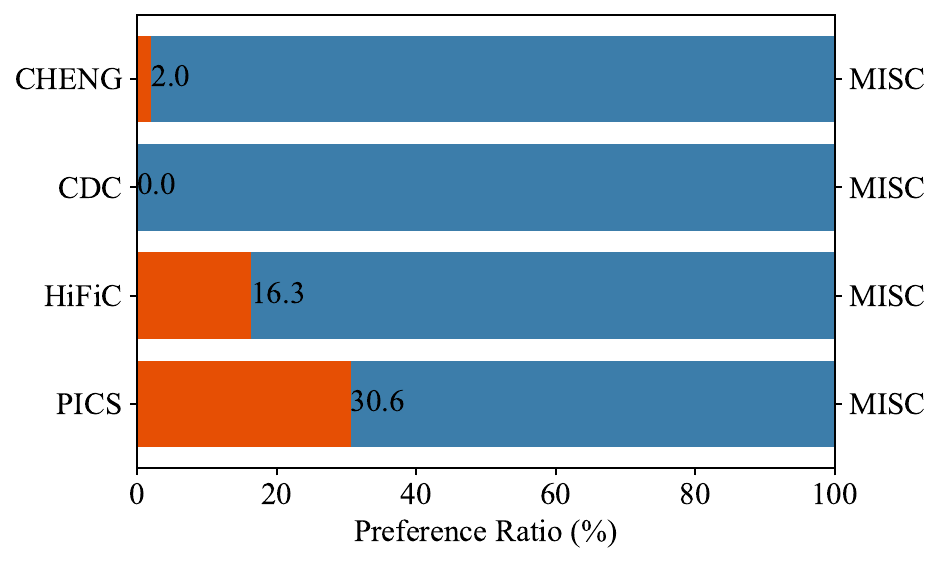}}
    \caption{Statistical results of user study in the NSI database CLIC-2020\cite{database:clic} and AIGI database AIGI-SCD. Humans subjectively believe that MISC is the best compression metric for both consistency and perception.}
    \label{fig:user}
\end{figure}

To verify the practicality of MISC in real-life scenarios, we conduct a subjective user study beyond the objective indicators, to analyze the human preference for the compressed image. We established an environment with standard lighting, displaying the ground truth centrally, and two compressed images on an iMac monitor with a resolution of up to 4,096 $\times$ 2,304 pixels. Viewers are required to select preferences between two images compressed by different algorithms, at both consistency and perception levels. The experiment involved 7 graduate students (4 males and 3 females) interacting with the interface in Fig. \ref{fig:html}.
The proposed MISC is compared with four state-of-art compression metrics, namely CHENG \cite{metric:nic-cheng}, CDC \cite{metric:diff-cdc}, HiFiC \cite{metric:gan-hific}, and PICS \cite{metric:diff-text}. A certain bitrate ($0.04\sim0.05$) is set for all those metrics for a fair comparison, by using MISC-3, and the bitrate in TABLE \ref{tab:main} for other metrics. The validation results illustrated in Fig. \ref{fig:user} demonstrate the superior performance of MISC across all evaluated criteria. Notably, MISC performs comparably to the PICS for consistency, and HiFiC for perception. Furthermore, compared to NSIs, AIGIs compressed by MISC were more preferred by human evaluators.
It is worth mentioning that this subjective result is slightly different from the objective indicators in TABLE \ref{tab:main}. For example, in AIGI compression, the HiFiC \cite{metric:gan-hific} achieves the best objective indicators, but Fig. \ref{fig:user} shows that MISC is more subjectively preferred. Therefore, appropriate objective perceptual quality assessment measures should be developed to inspire image compression at ultra-low bitrates.

\section{Conclusion}
In this paper, an image compression method called MISC is proposed at ultra-low bitrates. It can significantly reduce the storage space required and save bandwidth to transfer images. The framework of MISC consists of LMM/map/image encoders, and a general decoder. Experimental results on CLIC2020 and the AIGI-SCD database constructed in this paper show that MISC solves the trade-off problem between consistency and perception, as well as good scalability with bitrates. With the evolution of today's storage devices and communication protocols, we believe this LMM-driven methodology has the potential to facilitate a new paradigm for image compression.



\bibliographystyle{IEEEbib}
\bibliography{cite}

\begin{thebibliography}{10}

\bibitem{intro:tip-1}
Qi~Mao, Chongyu Wang, Meng Wang, Shiqi Wang, Ruijie Chen, Libiao Jin, and Siwei Ma,
\newblock ``Scalable face image coding via stylegan prior: Toward compression for human-machine collaborative vision,''
\newblock {\em IEEE Transactions on Image Processing}, vol. 33, pp. 408--422, 2024.

\bibitem{intro:tip-2}
Wenhong Duan, Zheng Chang, Chuanmin Jia, Shanshe Wang, Siwei Ma, Li~Song, and Wen Gao,
\newblock ``Learned image compression using cross-component attention mechanism,''
\newblock {\em IEEE Transactions on Image Processing}, vol. 32, pp. 5478--5493, 2023.

\bibitem{intro:tip-3}
Juan Wang, Yiping Duan, Xiaoming Tao, Mai Xu, and Jianhua Lu,
\newblock ``Semantic perceptual image compression with a laplacian pyramid of convolutional networks,''
\newblock {\em IEEE Transactions on Image Processing}, vol. 30, pp. 4225--4237, 2021.

\bibitem{my:5G}
Chunyi Li, Haoyang Li, Ning Yang, and Dazhi He,
\newblock ``A pbch reception algorithm in 5g broadcasting,''
\newblock in {\em IEEE International Symposium on Broadband Multimedia Systems and Broadcasting}, 2022.

\bibitem{his:6G}
Zicheng Zhang, Yingjie Zhou, Long Teng, Wei Sun, Chunyi Li, Xiongkuo Min, Xiao-Ping Zhang, and Guangtao Zhai,
\newblock ``Quality-of-experience evaluation for digital twins in 6g network environments,''
\newblock {\em IEEE Transactions on Broadcasting}, 2024.

\bibitem{intro:tradeoff-1}
Yochai Blau and Tomer Michaeli,
\newblock ``The perception-distortion tradeoff,''
\newblock in {\em Proceedings of the IEEE conference on computer vision and pattern recognition}, 2018, pp. 6228--6237.

\bibitem{intro:tradeoff-2}
Yochai Blau and Tomer Michaeli,
\newblock ``Rethinking lossy compression: The rate-distortion-perception tradeoff,''
\newblock in {\em International Conference on Machine Learning}. PMLR, 2019, pp. 675--685.

\bibitem{review:GIC}
Bolin Chen, Shanzhi Yin, Peilin Chen, Shiqi Wang, and Yan Ye,
\newblock ``Generative visual compression: A review,'' arXiv preprint arXiv:2402.02140, 2024.

\bibitem{metric:extreme}
Qi~Mao, Tinghan Yang, Yinuo Zhang, Zijian Wang, Meng Wang, Shiqi Wang, and Siwei Ma,
\newblock ``Extreme image compression using fine-tuned vqgans,'' arXiv preprint arXiv:2307.08265, 2023.

\bibitem{metric:Semantic}
Jianhui Chang, Jian Zhang, Jiguo Li, Shiqi Wang, Qi~Mao, Chuanmin Jia, Siwei Ma, and Wen Gao,
\newblock ``Semantic-aware visual decomposition for image coding,''
\newblock {\em International Journal of Computer Vision}, vol. 131, pp. 2333--2355, 2023.

\bibitem{intro:gpt4}
OpenAI,
\newblock ``Gpt-4 technical report,'' arXiv preprint arXiv:2303.08774, 2023.

\bibitem{intro:llama}
Hugo Touvron, Thibaut Lavril, Gautier Izacard, Xavier Martinet, Marie-Anne Lachaux, Timothée Lacroix, Baptiste Rozière, Naman Goyal, Eric Hambro, Faisal Azhar, Aurelien Rodriguez, Armand Joulin, Edouard Grave, and Guillaume Lample,
\newblock ``Llama: Open and efficient foundation language models,'' arXiv preprint arXiv:2302.13971, 2023.

\bibitem{gen:sd}
Robin Rombach, Andreas Blattmann, Dominik Lorenz, Patrick Esser, and Bj{\"o}rn Ommer,
\newblock ``High-resolution image synthesis with latent diffusion models,''
\newblock in {\em Proceedings of the IEEE/CVF conference on computer vision and pattern recognition}, 2022, pp. 10684--10695.

\bibitem{intro:xl}
Robin Rombach, Andreas Blattmann, and Björn Ommer,
\newblock ``Text-guided synthesis of artistic images with retrieval-augmented diffusion models,'' arXiv preprint arXiv:2207.13038, 2022.

\bibitem{intro:dalle}
Aditya Ramesh, Prafulla Dhariwal, Alex Nichol, Casey Chu, and Mark Chen,
\newblock ``Hierarchical text-conditional image generation with clip latents,'' arXiv preprint arXiv:2204.06125, 2022.

\bibitem{intro:aigc}
Jiayang Wu, Wensheng Gan, Zefeng Chen, Shicheng Wan, and Hong Lin,
\newblock ``Ai-generated content (aigc): A survey,'' arXiv preprint arXiv:2304.06632, 2023.

\bibitem{metric:tra-jpeg}
Athanassios~N. Skodras, Charilaos~A. Christopoulos, and Touradj Ebrahimi,
\newblock ``The jpeg 2000 still image compression standard,''
\newblock {\em IEEE Signal Process. Mag.}, vol. 18, pp. 36--58, 2001.

\bibitem{metric:tra-webp}
Oren Rippel and Lubomir Bourdev,
\newblock ``Real-time adaptive image compression,''
\newblock in {\em International Conference on Machine Learning}. PMLR, 2017, pp. 2922--2930.

\bibitem{metric:nic-mbt}
David~C. Minnen, Johannes Ball{\'e}, and George Toderici,
\newblock ``Joint autoregressive and hierarchical priors for learned image compression,''
\newblock in {\em Neural Information Processing Systems}, 2018.

\bibitem{metric:nic-rdoptq}
Junqi Shi, Ming Lu, and Zhan Ma,
\newblock ``Rate-distortion optimized post-training quantization for learned image compression,''
\newblock {\em IEEE Transactions on Circuits and Systems for Video Technology}, 2023.

\bibitem{metric:nic-srec}
Sheng Cao, Chao-Yuan Wu, and Philipp Krähenbühl,
\newblock ``Lossless image compression through super-resolution,'' arXiv preprint arXiv:2004.02872, 2020.

\bibitem{metric:nic-gao}
Wei Gao, Lvfang Tao, Linjie Zhou, Dinghao Yang, Xiaoyu Zhang, and Zixuan Guo,
\newblock ``Low-rate image compression with super-resolution learning,''
\newblock in {\em Proceedings of the IEEE/CVF Conference on Computer Vision and Pattern Recognition Workshops}, 2020, pp. 154--155.

\bibitem{metric:gan-gene}
Eirikur Agustsson, Michael Tschannen, Fabian Mentzer, Radu Timofte, and Luc~Van Gool,
\newblock ``Generative adversarial networks for extreme learned image compression,''
\newblock in {\em Proceedings of the IEEE/CVF International Conference on Computer Vision}, 2019, pp. 221--231.

\bibitem{metric:gan-ulc}
Fangyuan Gao, Xin Deng, Junpeng Jing, Xin Zou, and Mai Xu,
\newblock ``Extremely low bit-rate image compression via invertible image generation,''
\newblock {\em IEEE Transactions on Circuits and Systems for Video Technology}, 2023.

\bibitem{metric:gan-hific}
Fabian Mentzer, George~D Toderici, Michael Tschannen, and Eirikur Agustsson,
\newblock ``High-fidelity generative image compression,''
\newblock {\em Advances in Neural Information Processing Systems}, vol. 33, pp. 11913--11924, 2020.

\bibitem{metric:gan-multi}
Eirikur Agustsson, David Minnen, George Toderici, and Fabian Mentzer,
\newblock ``Multi-realism image compression with a conditional generator,''
\newblock in {\em Proceedings of the IEEE/CVF Conference on Computer Vision and Pattern Recognition}, 2023, pp. 22324--22333.

\bibitem{metric:gan-vari}
Shoma Iwai, Tomo Miyazaki, and Shinichiro Omachi,
\newblock ``Controlling rate, distortion, and realism: Towards a single comprehensive neural image compression model,''
\newblock in {\em Proceedings of the IEEE/CVF Winter Conference on Applications of Computer Vision}, 2024, pp. 2900--2909.

\bibitem{metric:diff-cdc}
Ruihan Yang and Stephan Mandt,
\newblock ``Lossy image compression with conditional diffusion models,'' arXiv preprint arXiv:2209.06950, 2022.

\bibitem{metric:diff-pan}
Zhihong Pan, Xin Zhou, and Hao Tian,
\newblock ``Extreme generative image compression by learning text embedding from diffusion models,'' arXiv preprint arXiv:2211.07793, 2022.

\bibitem{metric:diff-cmc}
Jiguo Li, Chuanmin Jia, Xinfeng Zhang, Siwei Ma, and Wen Gao,
\newblock ``Cross modal compression: Towards human-comprehensible semantic compression,''
\newblock in {\em ACM Multimedia}, 2021.

\bibitem{metric:diff-mcmc}
Junlong Gao, Chuanmin Jia, Zhimeng Huang, Shanshe Wang, Siwei Ma, and Wen Gao,
\newblock ``Rate-distortion optimized cross modal compression with multiple domains,''
\newblock {\em IEEE Transactions on Circuits and Systems for Video Technology}, pp. 1--1, 2024.

\bibitem{metric:diff-sgc}
Tom Bordin and Thomas Maugey,
\newblock ``Semantic based generative compression of images for extremely low bitrates,''
\newblock in {\em 2023 IEEE 25th International Workshop on Multimedia Signal Processing}. IEEE, 2023, pp. 1--6.

\bibitem{metric:diff-hfd}
Emiel Hoogeboom, Eirikur Agustsson, Fabian Mentzer, Luca Versari, George Toderici, and Lucas Theis,
\newblock ``High-fidelity image compression with score-based generative models,'' arXiv preprint arXiv:2305.18231, 2023.

\bibitem{metric:diff-text}
Eric Lei, Yiğit~Berkay Uslu, Hamed Hassani, and Shirin~Saeedi Bidokhti,
\newblock ``Text + sketch: Image compression at ultra low rates,'' arXiv preprint arXiv:2307.01944, 2023.

\bibitem{metric:tra-avc}
Thomas Wiegand, Gary~J. Sullivan, Gisle Bj{\o}ntegaard, and Ajay Luthra,
\newblock ``Overview of the h.264/avc video coding standard,''
\newblock {\em IEEE Trans. Circuits Syst. Video Technol.}, vol. 13, pp. 560--576, 2003.

\bibitem{metric:tra-hevc}
Gary~J. Sullivan, Jens-Rainer Ohm, Woojin Han, and Thomas Wiegand,
\newblock ``Overview of the high efficiency video coding (hevc) standard,''
\newblock {\em IEEE Transactions on Circuits and Systems for Video Technology}, vol. 22, pp. 1649--1668, 2012.

\bibitem{metric:tra-vvc}
Benjamin Bross, Ye-Kui Wang, Yan Ye, Shan Liu, Jianle Chen, Gary~J. Sullivan, and Jens-Rainer Ohm,
\newblock ``Overview of the versatile video coding (vvc) standard and its applications,''
\newblock {\em IEEE Transactions on Circuits and Systems for Video Technology}, vol. 31, pp. 3736--3764, 2021.

\bibitem{model:clip}
Alec Radford, Jong~Wook Kim, Chris Hallacy, Aditya Ramesh, Gabriel Goh, Sandhini Agarwal, Girish Sastry, Amanda Askell, Pamela Mishkin, Jack Clark, et~al.,
\newblock ``Learning transferable visual models from natural language supervision,''
\newblock in {\em International conference on machine learning}. PMLR, 2021, pp. 8748--8763.

\bibitem{database:kodak}
Gisle Bjontegaard,
\newblock ``Calculation of average psnr differences between rd-curves,''
\newblock {\em ITU SG16 Doc. VCEG-M33}, 2001.

\bibitem{database:div2k}
Eirikur Agustsson and Radu Timofte,
\newblock ``Ntire 2017 challenge on single image super-resolution: Dataset and study,''
\newblock in {\em Proceedings of the IEEE Conference on Computer Vision and Pattern Recognition Workshops}, July 2017.

\bibitem{database:clic}
Johannes Ball{\'e}, Philip~A Chou, David Minnen, Saurabh Singh, Nick Johnston, Eirikur Agustsson, Sung~Jin Hwang, and George Toderici,
\newblock ``Nonlinear transform coding,''
\newblock {\em IEEE Journal of Selected Topics in Signal Processing}, vol. 15, no. 2, pp. 339--353, 2020.

\bibitem{database:agiqa-3k}
Chunyi Li, Zicheng Zhang, Haoning Wu, Wei Sun, Xiongkuo Min, Xiaohong Liu, Guangtao Zhai, and Weisi Lin,
\newblock ``Agiqa-3k: An open database for ai-generated image quality assessment,''
\newblock {\em IEEE Transactions on Circuits and Systems for Video Technology}, 2023.

\bibitem{database:agiqa-1k}
Zicheng Zhang, Chunyi Li, Wei Sun, Xiaohong Liu, Xiongkuo Min, and Guangtao Zhai,
\newblock ``A perceptual quality assessment exploration for aigc images,''
\newblock in {\em IEEE International Conference on Multimedia and Expo Workshops}, 2023.

\bibitem{iqa:ssim}
Zhou Wang, Alan~C Bovik, Hamid~R Sheikh, and Eero~P Simoncelli,
\newblock ``Image quality assessment: from error visibility to structural similarity,''
\newblock {\em IEEE transactions on image processing}, vol. 13, no. 4, pp. 600--612, 2004.

\bibitem{iqa:fid}
Martin Heusel, Hubert Ramsauer, Thomas Unterthiner, Bernhard Nessler, and Sepp Hochreiter,
\newblock ``Gans trained by a two time-scale update rule converge to a local nash equilibrium,''
\newblock {\em Advances in neural information processing systems}, vol. 30, 2017.

\bibitem{add:q-bench}
Haoning Wu, Zicheng Zhang, Erli Zhang, Chaofeng Chen, Liang Liao, Annan Wang, Chunyi Li, Wenxiu Sun, Qiong Yan, Guangtao Zhai, et~al.,
\newblock ``Q-bench: A benchmark for general-purpose foundation models on low-level vision,'' arXiv preprint arXiv:2309.14181, 2023.

\bibitem{add:q-instruct}
Haoning Wu, Zicheng Zhang, Erli Zhang, Chaofeng Chen, Liang Liao, Annan Wang, Kaixin Xu, Chunyi Li, Jingwen Hou, Guangtao Zhai, et~al.,
\newblock ``Q-instruct: Improving low-level visual abilities for multi-modality foundation models,'' arXiv preprint arXiv:2311.06783, 2023.

\bibitem{add:q-boost}
Zicheng Zhang, Haoning Wu, Zhongpeng Ji, Chunyi Li, Erli Zhang, Wei Sun, Xiaohong Liu, Xiongkuo Min, Fengyu Sun, Shangling Jui, et~al.,
\newblock ``Q-boost: On visual quality assessment ability of low-level multi-modality foundation models,'' arXiv preprint arXiv:2312.15300, 2023.

\bibitem{add:q-align}
Haoning Wu, Zicheng Zhang, Weixia Zhang, Chaofeng Chen, Liang Liao, Chunyi Li, Yixuan Gao, Annan Wang, Erli Zhang, Wenxiu Sun, et~al.,
\newblock ``Q-align: Teaching lmms for visual scoring via discrete text-defined levels,'' arXiv preprint arXiv:2312.17090, 2023.

\bibitem{add:q-refine}
Chunyi Li, Haoning Wu, Zicheng Zhang, Hongkun Hao, Kaiwei Zhang, Lei Bai, Xiaohong Liu, Xiongkuo Min, Weisi Lin, and Guangtao Zhai,
\newblock ``Q-refine: A perceptual quality refiner for ai-generated image,'' arXiv preprint arXiv:2401.01117, 2024.

\bibitem{his:gms}
Zicheng Zhang, Wei Sun, Houning Wu, Yingjie Zhou, Chunyi Li, Xiongkuo Min, Guangtao Zhai, and Weisi Lin,
\newblock ``Gms-3dqa: Projection-based grid mini-patch sampling for 3d model quality assessment,'' arXiv preprint arXiv:2306.05658, 2023.

\bibitem{his:advancing}
Zicheng Zhang, Wei Sun, Yingjie Zhou, Haoning Wu, Chunyi Li, Xiongkuo Min, Xiaohong Liu, Guangtao Zhai, and Weisi Lin,
\newblock ``Advancing zero-shot digital human quality assessment through text-prompted evaluation,'' arXiv preprint arXiv:2307.02808, 2023.

\bibitem{my:aspect-qoe}
Chunyi Li, May Lim, Abdelhak Bentaleb, and Roger Zimmermann,
\newblock ``A real-time blind quality-of-experience assessment metric for http adaptive streaming,''
\newblock in {\em IEEE International Conference on Multimedia and Expo}, 2023.

\bibitem{my:xgc-vqa}
Xinhui Huang, Chunyi Li, Abdelhak Bentaleb, Roger Zimmermann, and Guangtao Zhai,
\newblock ``Xgc-vqa: A unified video quality assessment model for user, professionally, and occupationally-generated content,''
\newblock in {\em IEEE International Conference on Multimedia and Expo Workshops}, 2023.

\bibitem{my:cartoon}
Chunyi Li, Zicheng Zhang, Wei Sun, Xiongkuo Min, and Guangtao Zhai,
\newblock ``A full-reference quality assessment metric for cartoon images,''
\newblock in {\em IEEE 24th International Workshop on Multimedia Signal Processing}, 2022.

\bibitem{model:surgery}
Yi~Li, Hualiang Wang, Yiqun Duan, and Xiaomeng Li,
\newblock ``Clip surgery for better explainability with enhancement in open-vocabulary tasks,'' arXiv preprint arXiv:2304.05653, 2023.

\bibitem{model:diffbir}
Xinqi Lin, Jingwen He, Ziyan Chen, Zhaoyang Lyu, Ben Fei, Bo~Dai, Wanli Ouyang, Yu~Qiao, and Chao Dong,
\newblock ``Diffbir: Towards blind image restoration with generative diffusion prior,'' arXiv preprint arXiv:2308.15070, 2023.

\bibitem{quality:CLIPIQA}
Jianyi Wang, Kelvin~CK Chan, and Chen~Change Loy,
\newblock ``Exploring clip for assessing the look and feel of images,''
\newblock in {\em Proceedings of the AAAI Conference on Artificial Intelligence}, 2023, vol.~37, pp. 2555--2563.

\bibitem{quality:DBCNN}
Weixia Zhang, Kede Ma, Jia Yan, Dexiang Deng, and Zhou Wang,
\newblock ``Blind image quality assessment using a deep bilinear convolutional neural network,''
\newblock {\em IEEE Transactions on Circuits and Systems for Video Technology}, vol. 30, no. 1, pp. 36--47, 2018.

\bibitem{quality:LIQE}
Weixia Zhang, Guangtao Zhai, Ying Wei, Xiaokang Yang, and Kede Ma,
\newblock ``Blind image quality assessment via vision-language correspondence: A multitask learning perspective,''
\newblock in {\em Proceedings of the IEEE/CVF Conference on Computer Vision and Pattern Recognition}, 2023, pp. 14071--14081.

\bibitem{quality:NIQE}
Anish Mittal, Rajiv Soundararajan, and Alan~C Bovik,
\newblock ``Making a “completely blind” image quality analyzer,''
\newblock {\em IEEE Signal processing letters}, vol. 20, no. 3, pp. 209--212, 2012.

\bibitem{gen:turbo}
Axel Sauer, Dominik Lorenz, Andreas Blattmann, and Robin Rombach,
\newblock ``Adversarial diffusion distillation,'' arXiv preprint arXiv:2311.17042, 2023.

\bibitem{gen:MJ}
David Holz,
\newblock ``Midjourney,'' \url{https://www.midjourney.com}, 2023.

\bibitem{gen:ssd-1b}
Yatharth Gupta, Vishnu~V. Jaddipal, Harish Prabhala, Sayak Paul, and Patrick~Von Platen,
\newblock ``Progressive knowledge distillation of stable diffusion xl using layer level loss,'' arXiv preprint arXiv:2401.02677, 2024.

\bibitem{gen:Playground}
PlaygroundAI,
\newblock ``playground-v2-1024px-aesthetic,'' \url{https://playground.com}, 2023.

\bibitem{gen:dream}
dreamlike art,
\newblock ``dreamlike-photoreal-2.0,'' \url{https://dreamlike.art}, 2023.

\bibitem{gen:pixart}
Junsong Chen, Jincheng Yu, Chongjian Ge, Lewei Yao, Enze Xie, Yue Wu, Zhongdao Wang, James Kwok, Ping Luo, Huchuan Lu, and Zhenguo Li,
\newblock ``Pixart-$\alpha$: Fast training of diffusion transformer for photorealistic text-to-image synthesis,'' arXiv preprint arXiv:2310.00426, 2023.

\bibitem{gen:IF}
DeepFloyd,
\newblock ``If-i-xl-v1.0,'' \url{https://www.deepfloyd.ai}, 2023.

\bibitem{database:tecnick}
Nicola Asuni and Andrea Giachetti,
\newblock ``Testimages: a large-scale archive for testing visual devices and basic image processing algorithms.,''
\newblock in {\em Smart Tools and Applications in Graphics}, 2014, pp. 63--70.

\bibitem{add:define}
Vlad Hosu, Franz Hahn, Mohsen Jenadeleh, Hanhe Lin, Hui Men, Tam{\'a}s Szir{\'a}nyi, Shujun Li, and Dietmar Saupe,
\newblock ``The konstanz natural video database (konvid-1k),''
\newblock in {\em 2017 Ninth international conference on quality of multimedia experience}. IEEE, 2017, pp. 1--6.

\bibitem{iqa:lpips}
Richard Zhang, Phillip Isola, Alexei~A. Efros, Eli Shechtman, and Oliver Wang,
\newblock ``The unreasonable effectiveness of deep features as a perceptual metric,'' arXiv preprint arXiv:1801.03924, 2018.

\bibitem{metric:nic-bmshj}
Johannes Ball{\'e}, David Minnen, Saurabh Singh, Sung~Jin Hwang, and Nick Johnston,
\newblock ``Variational image compression with a scale hyperprior,''
\newblock in {\em International Conference on Learning Representations}, 2018.

\bibitem{metric:nic-cheng}
Zhengxue Cheng, Heming Sun, Masaru Takeuchi, and J.~Katto,
\newblock ``Learned image compression with discretized gaussian mixture likelihoods and attention modules,''
\newblock {\em 2020 IEEE/CVF Conference on Computer Vision and Pattern Recognition}, pp. 7936--7945, 2020.

\bibitem{quality:brisque}
Anish Mittal, Anush~Krishna Moorthy, and Alan~Conrad Bovik,
\newblock ``No-reference image quality assessment in the spatial domain,''
\newblock {\em IEEE Transactions on image processing}, vol. 21, no. 12, pp. 4695--4708, 2012.

\bibitem{quality:HyperIQA}
Shaolin Su, Qingsen Yan, Yu~Zhu, Cheng Zhang, Xin Ge, Jinqiu Sun, and Yanning Zhang,
\newblock ``Blindly assess image quality in the wild guided by a self-adaptive hyper network,''
\newblock in {\em Proceedings of the IEEE/CVF Conference on Computer Vision and Pattern Recognition}, 2020, pp. 3667--3676.

\bibitem{database:DiffusionDB}
Zijie~J Wang, Evan Montoya, David Munechika, Haoyang Yang, Benjamin Hoover, and Duen~Horng Chau,
\newblock ``Diffusiondb: A large-scale prompt gallery dataset for text-to-image generative models,''
\newblock {\em arXiv preprint arXiv:2210.14896}, 2022.

\bibitem{database:GenImage}
Mingjian Zhu, Hanting Chen, Qiangyu Yan, Xudong Huang, Guanyu Lin, Wei Li, Zhijun Tu, Hailin Hu, Jie Hu, and Yunhe Wang,
\newblock ``Genimage: A million-scale benchmark for detecting ai-generated image,'' arXiv preprint arXiv:2306.08571, 2023.

\bibitem{database:pokemon}
Justin N.~M. Pinkney,
\newblock ``Pokemon blip captions,'' \url{https://huggingface.co/datasets/lambdalabs/pokemon-blip-captions/}, 2022.

\bibitem{other:adam}
Diederik~P Kingma and Jimmy Ba,
\newblock ``Adam: A method for stochastic optimization,'' arXiv preprint arXiv:1412.6980, 2014.

\end{thebibliography}

\end{document}